\documentclass[10pt,twocolumn,letterpaper]{article}

\usepackage[pagenumbers]{cvpr} 

\usepackage{graphicx}
\usepackage{booktabs}
\usepackage{amsmath}
\usepackage{amssymb}
\usepackage{siunitx}
\usepackage{hhline}
\usepackage{subcaption}
\usepackage{csquotes}
\usepackage{multirow}
\usepackage{pifont}
\usepackage[super]{nth}
\usepackage{mathtools,bbm}
\usepackage{bm}
\usepackage{textcomp}
\usepackage{xcolor}
\usepackage{colortbl}
\usepackage[accsupp]{axessibility}  

\newcolumntype{L}[1]{>{\raggedright\let\newline\\\arraybackslash\hspace{0pt}}m{#1}}
\newcolumntype{C}[1]{>{\centering\let\newline\\\arraybackslash\hspace{0pt}}m{#1}}
\newcolumntype{R}[1]{>{\raggedleft\let\newline\\\arraybackslash\hspace{0pt}}m{#1}}

\definecolor{darkergreen}{RGB}{21, 152, 56}
\definecolor{red2}{RGB}{252, 54, 65}
\newcommand{\cmark}{\textcolor{darkergreen}{\ding{51}}}
\newcommand{\xmark}{\textcolor{red2}{\ding{55}}}

\definecolor{cvprblue}{rgb}{0.21,0.49,0.74}
\usepackage[pagebackref,breaklinks,colorlinks,allcolors=cvprblue]{hyperref}

\def\paperID{31897} 
\def\confName{CVPR}
\def\confYear{2026}

\title{Zero-Shot Reconstruction of Animatable 3D Avatars with Cloth Dynamics \\from a Single Image}

\author{
    Joohyun Kwon \qquad
    Geonhee Sim \qquad
    Gyeongsik Moon \\
    Korea University \\
    {\tt\small \{juheanqueen, kh6362, mks0601\}@korea.ac.kr} \\
    {\small \url{https://juhyeon-kwon.github.io/DynaAvatar.github.io/}}
}

\begin{document}

\twocolumn[{
\maketitle
\vspace{-2.5em}
\centerline{
\includegraphics[width=\linewidth,trim={4pt 4pt 4pt 4pt}]{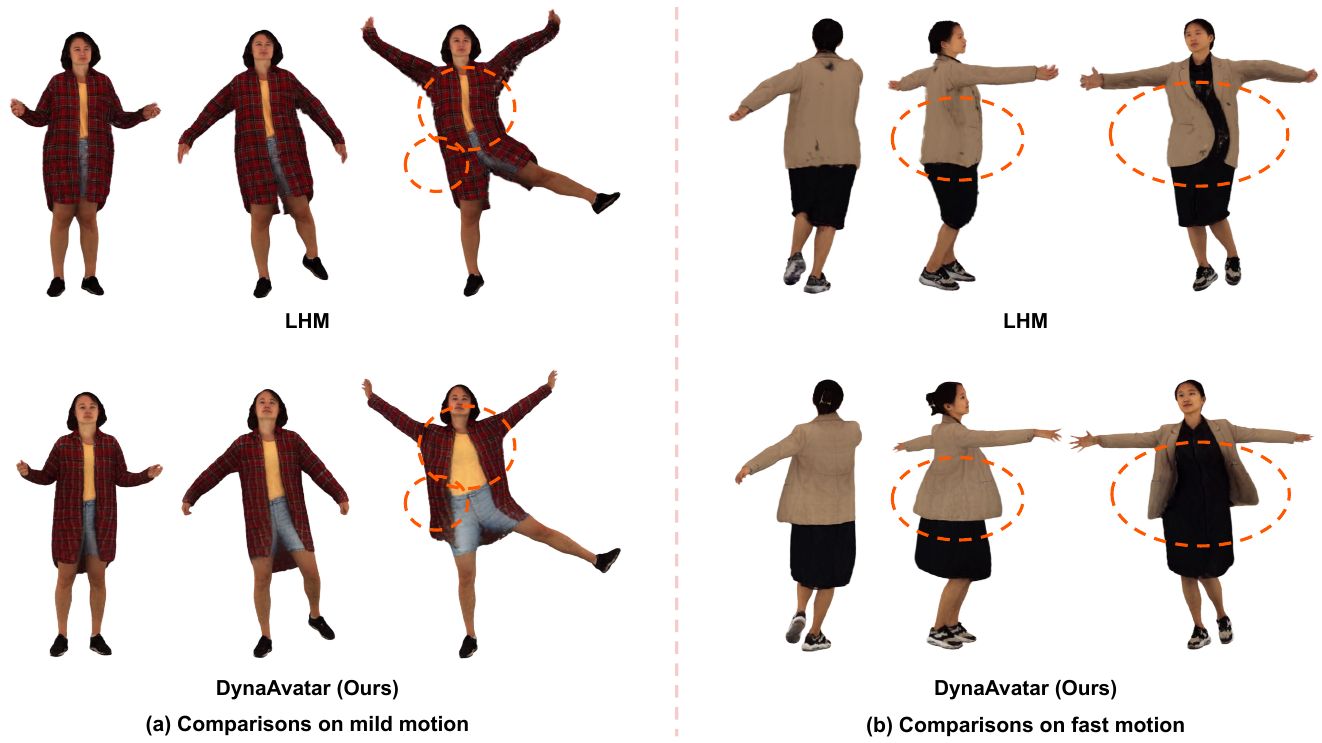}}
\vspace{-0.5em}
\captionof{figure}
{
Comparison between LHM~\cite{qiu2025LHM} and our DynaAvatar on both mild and fast motions.
Unlike prior single-image-based methods, DynaAvatar can reconstruct animatable 3D human avatars that exhibit motion-dependent cloth dynamics.
}
\label{fig:intro_compare}
\vspace{1em}
}]

\maketitle

\begin{abstract}
Existing single-image 3D human avatar methods primarily rely on rigid joint transformations, limiting their ability to model realistic cloth dynamics.
We present \textbf{DynaAvatar}, a zero-shot framework that reconstructs animatable 3D human avatars with motion-dependent cloth dynamics from a single image.
Trained on large-scale multi-person motion datasets, DynaAvatar employs a Transformer-based feed-forward architecture that directly predicts dynamic 3D Gaussian deformations without subject-specific optimization.
To overcome the scarcity of dynamic captures, we introduce a static-to-dynamic knowledge transfer strategy: a Transformer pretrained on large-scale static captures provides strong geometric and appearance priors, which are efficiently adapted to motion-dependent deformations through lightweight LoRA fine-tuning on dynamic captures.
We further propose the DynaFlow loss, an optical flow–guided objective that provides reliable motion-direction geometric cues for cloth dynamics in rendered space.
Finally, we reannotate the missing or noisy SMPL-X fittings in existing dynamic capture datasets, as most public dynamic capture datasets contain incomplete or unreliable fittings that are unsuitable for training high-quality 3D avatar reconstruction models.
Experiments demonstrate that DynaAvatar produces visually rich and generalizable animations, outperforming prior methods.
\end{abstract}

\section{Introduction}

Creating realistic and animatable 3D human avatars from monocular inputs has long been a fundamental goal in computer vision, graphics, and virtual human research.
Most existing single-image animatable avatar reconstruction methods~\cite{qiu2025anigs,qiu2025LHM,zhuang2025idol} are limited to skeletal-based deformation, where the human body is animated primarily by rigid transformations of body joints.
While effective for body articulation, such representations inherently lack the ability to model non-rigid cloth dynamics, resulting in over-rigid motion and a loss of visual realism during animation.

Several personalized avatar methods~\cite{peng2021neural,peng2021animatable,kwon2021neural,bagautdinov2021driving,zheng2023avatarrex,li2024animatable,moreau2024human,xu2025sequential} reconstruct dynamic human models from multi-view videos of individual subjects.
Although these approaches capture subject-specific geometry and clothing deformation, they require a separate capture and optimization process for each person, making them impractical to apply to arbitrary subjects.
This dependency on multi-view capture severely limits their scalability and usability, especially when animatable avatars are needed for new individuals without dedicated data collection.

In this work, we present \textbf{DynaAvatar}, a novel framework that generates animatable 3D avatars with motion-dependent cloth dynamics from a single image in a zero-shot manner.
We define zero-shot as the ability to reconstruct avatars for unseen identities without any subject-specific fine-tuning or optimization.
Unlike personalized approaches~\cite{peng2021neural,peng2021animatable,kwon2021neural,bagautdinov2021driving,zheng2023avatarrex,li2024animatable,moreau2024human,xu2025sequential,moon2024expressive,sim2025persona,qiu2025anigs}, which fit a model to each individual, DynaAvatar learns motion-dependent deformation priors that generalize across identities, enabling feed-forward avatar reconstruction.
Our Transformer~\cite{vaswani2017attention,esser2024scaling}-based feed-forward architecture directly predicts dynamic 3D Gaussian deformations without any subject-specific optimization, allowing fast and scalable avatar reconstruction.

To address the limited availability of large-scale dynamic captures covering diverse clothing and motions, we introduce a static-to-dynamic knowledge transfer strategy that leverages pretrained static representations for dynamic learning.
We adopt a Transformer pretrained on large-scale static captures to provide strong geometric and appearance priors of human bodies and garments.
During dynamic training, we incorporate a Dynamic Transformer trained from scratch to specialize in motion-dependent deformation modeling, while efficiently adapting the pretrained Static Transformer using lightweight LoRA adapters~\cite{hu2022lora}.
This transfer-based design enables DynaAvatar to inherit rich static knowledge for geometry and appearance, while effectively learning motion-aware dynamic behaviors even with limited dynamic supervision.

To further enhance the learning of large and complex cloth movements, we introduce the DynaFlow loss function, an optical flow–guided correspondence loss that provides explicit pixel-wise motion-direction cues in the rendered screen space.
Unlike conventional image-space reconstruction losses that rely on local pixel similarity, DynaFlow leverages optical flow to establish reliable correspondences even under fast or large non-rigid cloth deformations, while providing geometry-only supervision that avoids the color–geometry ambiguity inherent to image losses. This leads to a more accurate modeling of motion-dependent cloth dynamics.

\begin{table}[t]
\footnotesize
\centering
\setlength\tabcolsep{1.0pt}
\def\arraystretch{1.1}
\caption{
Comparison of existing avatar reconstruction methods and the proposed DynaAvatar.
Each column indicates whether the method reconstructs avatars in a zero-shot manner (\textit{i.e.}, without subject-specific optimization), whether it supports cloth dynamics, and whether it operates from a single input image.
}
\vspace*{-3mm}
\begin{tabular}{L{2.5cm}|C{1.4cm}|C{1.9cm}|C{1.8cm}}
\specialrule{.1em}{.05em}{.05em}
Methods & Zero shot & Cloth dynamics & Single image \\ \hline
ExAvatar~\cite{moon2024expressive} & \xmark & \xmark & \xmark \\ 
IDOL~\cite{zhuang2025idol} & \cmark & \xmark & \cmark \\ 
AniGS~\cite{qiu2025anigs} & \xmark & \xmark & \cmark \\ 
LHM~\cite{qiu2025LHM} & \cmark & \xmark & \cmark \\ 
PERSONA~\cite{sim2025persona} & \xmark & \cmark & \cmark \\
MPMAvatar~\cite{lee2025mpmavatar} & \xmark & \cmark & \xmark \\
SeqAvatar~\cite{xu2025sequential} & \xmark & \cmark & \xmark \\
\textbf{DynaAvatar (Ours)} & \cmark & \cmark & \cmark \\
\specialrule{.1em}{.05em}{.05em}
\end{tabular}
\vspace*{-5mm}
\label{table:compare_novelty}
\end{table}

Finally, we reannotate existing dynamic capture datasets~\cite{cheng2023dna,wang20244d,icsik2023humanrf} with accurate SMPL-X~\cite{pavlakos2019expressive} fittings.
Although existing dynamic capture datasets provide multi-view videos with diverse clothing and motions, they often include missing or highly noisy SMPL-X parameters, making them unsuitable for training high-quality 3D avatar reconstruction models.
Our reannotation yields accurate and complete SMPL-X parameters, enabling over 11M high-quality image supervisions.

By integrating the static-to-dynamic knowledge transfer, DynaFlow loss function, and reannotated fittings, DynaAvatar effectively captures realistic and temporally coherent cloth dynamics across diverse motions.
Extensive experiments confirm that our framework significantly improves the realism and generalization of single-image 3D avatar generation, bridging the gap between static reconstruction and dynamic animation.

Our main contributions are as follows:
\begin{itemize}
\item We propose DynaAvatar, the first zero-shot framework that reconstructs animatable 3D avatars with motion-dependent cloth dynamics from a single image.

\item We introduce a static-to-dynamic knowledge transfer strategy that leverages pretrained static representations for geometry and appearance, and fine-tunes them with lightweight LoRA adaptation to learn motion-dependent deformations.

\item We propose the DynaFlow loss function, an optical flow–guided correspondence loss that provides pixel-wise motion-direction cues for more accurate supervision of large and complex cloth dynamics, offering geometry-only guidance that avoids the color–geometry ambiguity of image losses.

\item We reannotate existing dynamic capture datasets to curate a refined dataset with accurate SMPL-X parameters, providing consistent and reliable supervision for dynamic avatar training.
\end{itemize}

\section{Related works}

\noindent\textbf{Zero-shot 3D animatable avatars from a single image.}
Zero-shot 3D animatable avatar reconstruction methods aim to reconstruct avatars in a feed-forward manner without any subject-specific optimization, by being trained on large-scale datasets.
Since they eliminate the need for per-subject optimization, they achieve significantly shorter inference times compared to methods that rely on personalized fitting.
IDOL~\cite{zhuang2025idol} is trained on a large-scale generated dataset to reconstruct 3D Gaussian avatars from a single image.
LHM~\cite{qiu2025LHM} follows the spirit of IDOL by adopting a multimodal Transformer~\cite{vaswani2017attention,esser2024scaling} architecture for zero-shot 3D human reconstruction, while PF-LHM~\cite{qiu2025pf} further explores pose-free multi-image inputs for improved reconstruction robustness.
However, all of the above methods have limited animation fidelity, as they primarily rely on rigid body joint transformations for animation.
In contrast, our DynaAvatar is a zero-shot framework that can effectively model motion-dependent cloth dynamics, enabling more realistic and expressive avatar animations.

\noindent\textbf{3D animatable avatars with subject-specific optimizations.}
Peng~\etal~\cite{peng2021neural,peng2021animatable} and Zheng~\etal~\cite{zheng2023avatarrex} optimize personalized NeRF~\cite{mildenhall2021nerf}-based avatars using multi-view videos of each individual.
GaussianAvatar~\cite{hu2023gaussianavatar} and ExAvatar~\cite{moon2024expressive} instead optimize personalized 3DGS~\cite{kerbl20233d}-based avatars from monocular videos.
Dyco~\cite{chen2024within} introduce delta pose
sequence to effectively model temporal appearance variations.
AniGS~\cite{qiu2025anigs} synthesizes multi-view images and optimizes a 3DGS-based avatar on them, while PERSONA~\cite{sim2025persona} generates pose-rich videos for optimization.
SeqAvatar~\cite{xu2025sequential}, MPMAvatar~\cite{lee2025mpmavatar}, and Zhan~\etal~\cite{zhan2025real} performs optimization on multi-view videos.
All of the above methods rely on subject-specific optimization, making them unsuitable for arbitrary individuals when video captures are unavailable. 
Although some methods~\cite{qiu2025anigs,sim2025persona} support single-image optimization, they still suffer from slow inference compared to feed-forward architectures.

\noindent\textbf{Cloth dynamics.}
Early works~\cite{habermann2020deepcap,guo2024reloo,qiu2023rec} focused on 3D clothing reconstruction without animation capability.
More recent studies aim to recover simulation-ready 3D garment representations, where physics-based simulation is used to animate the reconstructed garments.
HOOD~\cite{grigorev2023hood} leverages graph neural networks, multi-level message passing, and unsupervised training to predict realistic clothing dynamics, while ContourCraft~\cite{grigorev2024contourcraft}, built upon HOOD, further resolves garment interpenetrations.
GaussianGarment~\cite{rong2025gaussian} represents clothing using a hybrid of 3D meshes and Gaussian textures, and PGC~\cite{guo2025pgc} reconstructs simulation-ready garments from multi-view images of a static human pose.
AIpparel~\cite{nakayama2025aipparel} and ChatGarment~\cite{bian2025chatgarment} enable single-image garment recovery by predicting simulation-ready clothing representations such as sewing patterns or parametric meshes.

Despite their progress, these physics-driven or simulation-based approaches face several limitations.
They either require clean cloth meshes or calibrated multi-view captures, which are difficult to obtain in real-world settings, or rely on automatically estimated 3D poses that are often inaccurate for in-the-wild single images or monocular videos.
Such imperfect or penetrating poses frequently cause instability in physics-based systems, resulting in cloth drifting and physically implausible deformations.
In contrast, our DynaAvatar learns data-driven, motion-dependent deformations without relying on explicit physical constraints, maintaining stability even under imperfect pose conditions.
Moreover, our framework reconstructs a full-body photorealistic avatar, whereas prior methods focus solely on clothing geometry without modeling the entire human body.

\noindent\textbf{Generative human animation.}
With the rapid progress of image and video generative models~\cite{ho2020denoising,blattmann2023stable}, many approaches animate a person from a single reference image using 2D pose sequences.
These methods directly generate human videos without explicit 3D representations. Diffusion-based frameworks~\cite{xu2024magicanimate,hu2024animate,zhang2025mimicmotion} improve temporal coherence and pose control, while others~\cite{men2025mimo,tu2025stableanimator} enhance spatial consistency and identity preservation. SMPL-conditioned models~\cite{zhu2024champ,shao2024360,hu2025humangif} enable controllable motion, and additional works extend facial motion or implicit motion modeling~\cite{luo2025dreamactor,song2025x}.
Although these models produce photorealistic results, they remain tied to their training setup: inputs and poses must be pixel-aligned, subjects must stay near the image center, and outputs have fixed resolution.
In contrast, DynaAvatar builds upon 3DGS~\cite{kerbl20233d}, supporting arbitrary poses, spatial layouts, and scalable resolutions with consistent 3D geometry and appearance.

\begin{figure*}[t]
\begin{center}
\includegraphics[width=\linewidth]{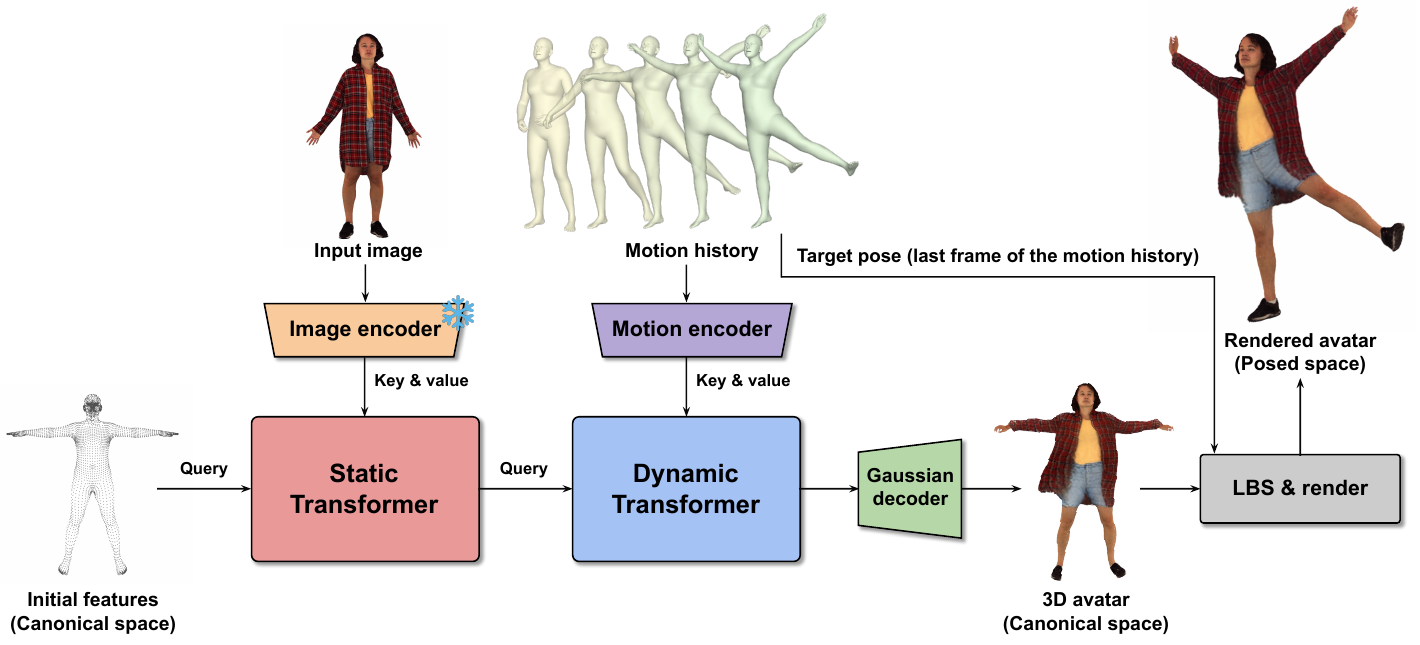}
\end{center}
\vspace*{-6mm}
\caption{
Overall pipeline of the proposed DynaAvatar.
We first extract detailed geometry and appearance without cloth dynamics using a Static Transformer.
Next, cloth dynamics are incorporated from motion history through a Dynamic Transformer.
The final 3D avatar in canonical space is reconstructed using a Gaussian decoder and then animated and rendered with LBS and a 3DGS renderer.
Since the canonical avatar already encodes motion-dependent cloth dynamics, the animation produced by LBS faithfully maintains these dynamics.
}
\vspace*{-3mm}
\label{fig:pipeline}
\end{figure*}

\section{DynaAvatar}

\subsection{Pipeline}

Fig.~\ref{fig:pipeline} illustrates the overall pipeline of the proposed DynaAvatar.
Given a single image of an arbitrary person, DynaAvatar extracts features using two Transformer modules~\cite{vaswani2017attention,esser2024scaling}.

\noindent\textbf{Static Transformer.}
The Static Transformer captures detailed geometry and appearance without modeling cloth dynamics.
It takes image tokens $\mathbf{T}_\text{I}$ extracted from the input image via pretrained encoders (Sapiens~\cite{khirodkar2024sapiens} and DINOv2~\cite{oquab2023dinov2}) as keys and values, while initial 3D point tokens $\mathbf{T}_\text{3D}$ (\emph{i.e.}, positional encoding~\cite{mildenhall2021nerf} of SMPL-X template vertices), serve as queries.
We freeze the pre-trained image encoder.
The 3D point tokens are then updated using a Multimodal Transformer Block (MM)~\cite{esser2024scaling}: $\mathbf{T}_\text{3D}, \mathbf{T}_\text{I} \leftarrow \text{MM}(\mathbf{T}_\text{3D}, \mathbf{T}_\text{I}; \mathbf{F}_\text{I})$.
The global context feature $\mathbf{F}_\text{I}$, obtained by averaging Sapiens image tokens, is used for  modulation via Adaptive Layer Normalization (AdaLN).

\noindent\textbf{Motion encoder.}
A motion encoder processes the motion history covering one second (15 time steps) to produce a motion tokens $\mathbf{T}_\text{M}$.
The motion history includes 3D poses, 3D pose velocities, 3D pose accelerations, and 3D keypoint velocities, where each pose is represented using the 6D rotation parameterization~\cite{zhou2019continuity}.
If the motion history is unavailable (\emph{e.g.}, the first frame of a video or a single-image demonstration), all past motions—excluding the current frame—are initialized to zero.
To ensure consistency, the motion history is transformed into a canonical world coordinate system, as the up-vector of each 3D pose may vary.
When world axes are provided by the dataset, they are used for alignment; otherwise, the camera’s $y$-axis is assumed as the up-vector.
This normalization prevents inconsistencies in motion semantics across poses.
The motion encoder consists of positional encoding followed by several multi-layer perceptrons (MLPs).

\noindent\textbf{Dynamic Transformer.}
The Dynamic Transformer refines the static features by integrating motion-aware cloth dynamics.
The motion tokens $\mathbf{T}_\text{M}$ produced by the motion encoder, together with the Static Transformer output $\mathbf{T}_\text{3D}$, are fed into the Dynamic Transformer.
Specifically, $\mathbf{T}_\text{M}$ acts as keys and values, while $\mathbf{T}_\text{3D}$ serves as queries within its MM~\cite{esser2024scaling}:  $\mathbf{T}_\text{3D}, \mathbf{T}_\text{M} \leftarrow \text{MM}(\mathbf{T}_\text{3D}, \mathbf{T}_\text{M}; \mathbf{F}_\text{M})$. 
The pose feature $\mathbf{F}_\text{M}$ is defined as the last element of $\mathbf{T}_\text{M}$ and serves as the condition for AdaLN.

\noindent\textbf{Gaussian Decoder.}
The output features from the Dynamic Transformer are fed into a Gaussian decoder, which reconstructs 3DGS~\cite{kerbl20233d} representations (\emph{i.e.}, mean, scale, rotation, opacity, and color) to form the final avatar in the canonical space with motion-dependent cloth dynamics.
In addition, it predicts skinning weight offsets, enabling each Gaussian to adapt its animation based on the motion history.
The decoder itself is implemented as a single linear layer.

\noindent\textbf{Animation and Rendering.}
Finally, the canonical avatar is animated using LBS with refined skinning weights, which combine the predicted offsets with diffused skinning weights~\cite{qiu2023rec,sim2025persona}, and rendered using a 3DGS renderer.
Note that the canonical avatar already encodes motion-dependent cloth dynamics, while the refined skinning weights further enhance these dynamics; thus, applying LBS naturally preserves realistic motion during animation.

\subsection{Static-to-dynamic knowledge transfer}

Learning realistic motion-dependent cloth dynamics ideally requires large-scale dynamic capture datasets, which are extremely costly to collect due to multi-view synchronization, temporal calibration, and garment diversity requirements~\cite{cheng2023dna,wang20244d,xiong2024mvhumannet,icsik2023humanrf}.
In contrast, static captures~\cite{zhuang2025idol,han2023high,yu2021function4d,liu2016deepfashion,renderpeople}
are far more abundant and provide high-quality geometry and appearance supervision,
though they lack temporal deformation cues.

Following the recent paradigm of leveraging large pretrained models for knowledge transfer~\cite{xu2024magicanimate,hu2024animate,zhang2025mimicmotion,men2025mimo,tu2025stableanimator}, we adopt a Transformer pretrained on large-scale static captures~\cite{qiu2025LHM} as the static backbone of DynaAvatar.
This pretrained model provides strong geometric and appearance priors that facilitate the subsequent learning of motion-dependent dynamics.
During dynamic training, we introduce a Dynamic Transformer trained from scratch while fine-tuning the pretrained Static Transformer using lightweight LoRA adapters~\cite{hu2022lora}.
This transfer-based adaptation allows DynaAvatar to effectively learn realistic cloth dynamics
even with limited dynamic supervision, benefiting from knowledge distilled from large-scale static data.

\begin{figure}[t]
\begin{center}
\includegraphics[width=0.9\linewidth]{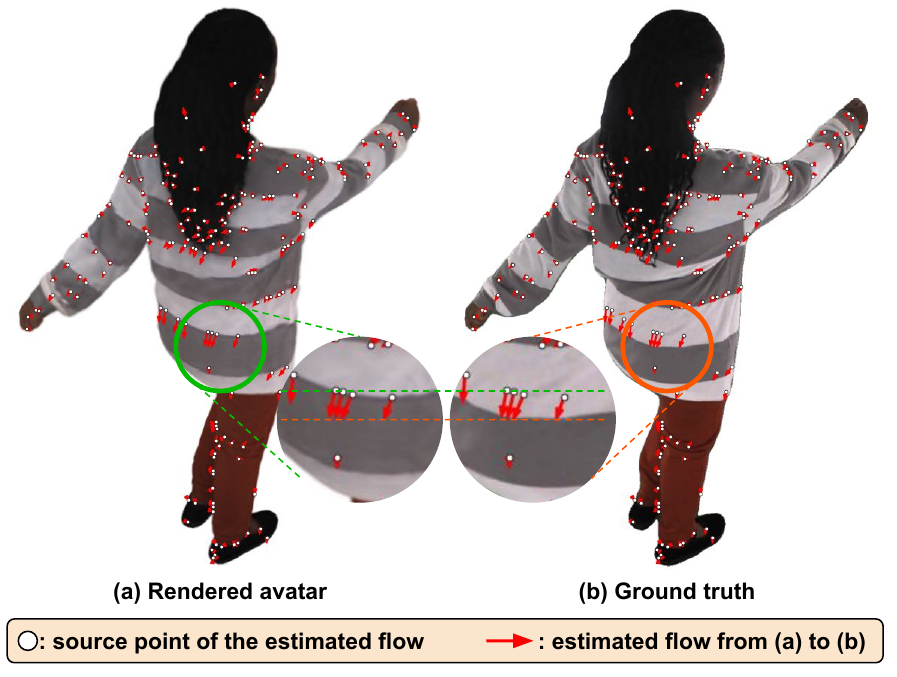}
\end{center}
\vspace*{-6mm}
\caption{
Visualization of the proposed DynaFlow loss.
Our DynaFlow loss encourages the Gaussians at source locations (black-outlined white circles) to move toward the endpoints of the estimated flow vectors.
}
\vspace*{-3mm}
\label{fig:dynaflow_loss}
\end{figure}

\subsection{DynaFlow loss function}

Image reconstruction losses entangle geometry and color, creating ambiguity that weakens geometric supervision. They also operate on local patches, making them ineffective for strong cloth deformations that require long-range correspondence. To address these limitations, we introduce DynaFlow loss function $\mathcal{L}_\text{flow}$, a geometry-only, flow-based supervision that provides explicit, deformation-aligned correspondence cues. By decoupling geometry from appearance, DynaFlow supplies structural signals that image losses cannot capture, enabling accurate modeling of large and complex cloth motions.
Fig.~\ref{fig:dynaflow_loss} illustrates our DynaFlow loss, which leverages optical flow to establish correspondences between the rendered and ground-truth images even when deformation exceeds the receptive field of conventional losses. 

To extract robust geometric cues, we render not only an RGB image but also an $xy$ map $\mathbf{M} \in \mathbb{R}^{H \times W \times 2}$ substituting Gaussian colors with their projected screen-space $xy$ coordinates.
We compute optical flow between the rendered and ground truth images using LightGlue~\cite{lindenberger2023lightglue}, obtaining $N$ matched source and target pixel coordinates, $\mathbf{p}_\text{src}$ and $\mathbf{p}_\text{tgt}$, respectively. 
By grid-sampling the $xy$ map $\mathbf{M}$ at $\mathbf{p}_\text{src}$ and enforcing the sampled coordinates to match the flow targets $\mathbf{p}_\text{tgt}$, DynaFlow injects direct pixel-level displacement supervision, defined as: $\mathcal{L}_\text{flow} = \frac{1}{N}\sum\|\mathbf{M}(\mathbf{p}_\text{src})-\mathbf{p}_\text{tgt}\|_1$.
This design allows gradients to backpropagate through $\mathbf{M}$, thereby directly correcting the 2D positions of the Gaussians.
Our Dynaflow loss resolves the geometry–color entanglement that causes image losses to blur or smooth out large cloth motions, allowing our model to reconstruct sharper boundaries and more faithful dynamic deformations. 
We cap the number of matches $N$ at 1024 for stability, and LightGlue’s efficiency keeps the additional training cost minimal.
The $\mathcal{L}_\text{flow}$ is activated only after the midpoint of training, since early-stage renderings are inaccurate and produce unreliable optical flow.

In addition to DynaFlow, we supervise the rendered avatars using a combination of $L1$, SSIM, mask, and LPIPS~\cite{zhang2018unreasonable} losses. We further adopt geometry regularizers from prior works~\cite{moon2024expressive,sim2025persona}, including the Laplacian regularizer, to stabilize face and hand geometry—regions that occupy only a small portion of the image and thus receive weak direct supervision.

\begin{figure}[t]
\begin{center}
\includegraphics[width=\linewidth]{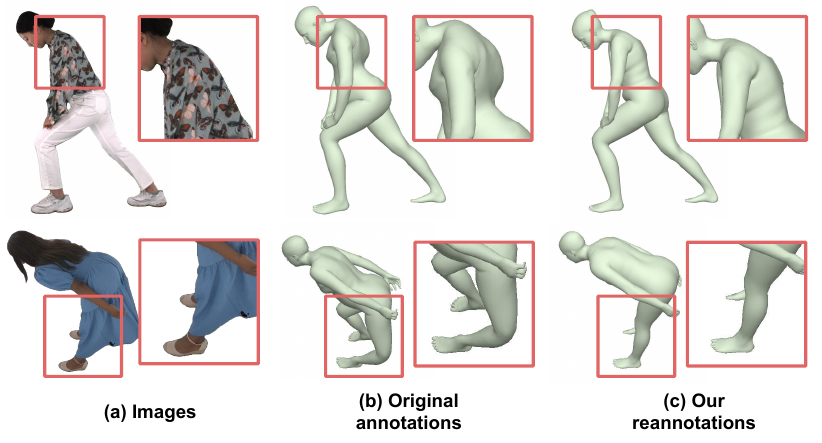}
\end{center}
\vspace*{-6mm}
\caption{
Comparison between (b) the original annotations and (c) our reannotations for the DNA-Rendering~\cite{cheng2023dna} (top) and Actors-HQ~\cite{icsik2023humanrf} (bottom) datasets.
}
\vspace*{-3mm}
\label{fig:reannotations}
\end{figure}

\section{Reannotating dynamic capture datasets}\label{sec:reannotate}

Fig.~\ref{fig:reannotations} compares (b) existing SMPL-X fittings and (c) our reannotated results.
Our reannotation yields accurate and complete SMPL-X parameters, enabling over 11M high-quality image supervisions.
Although multiple dynamic capture datasets have been introduced~\cite{cheng2023dna,wang20244d,xiong2024mvhumannet,icsik2023humanrf}, most are not directly suitable for training or evaluating DynaAvatar due to incomplete or noisy SMPL-X annotations.
DNA-Rendering~\cite{cheng2023dna} provides SMPL-X fittings for only about 20\% of frames, and even those often exhibit significant errors and temporal jitter.
MVHumanNet~\cite{xiong2024mvhumannet} offers downsampled sequences at only 5 frames per second, which is insufficient for modeling motion-dependent cloth dynamics.
4D-Dress~\cite{wang20244d} provides only gender-specific fittings without gender-neutral counterparts, while most 3D avatar reconstruction methods~\cite{qiu2025LHM,zhuang2025idol} (including ours) require gender-neutral models for consistency.
Actors-HQ~\cite{icsik2023humanrf} also contains inaccurate SMPL-X fittings.
Such inaccuracies lead to pixel misalignment between the rendered outputs and the target images, severely degrading supervision quality, as the primary image reconstruction loss assumes approximate pixel alignment between the rendered avatar and the ground-truth image.

To address the missing or noisy original annotations, we reconstruct refined SMPL-X fittings for all frames using a unified reannotation pipeline.
We first predict 2D whole-body keypoints for all images using DWPose~\cite{yang2023effective} and initialize SMPL-X parameters from the frontal-view image via SMLest-X~\cite{yin2025smplest}.
The 3D translation is triangulated from 2D keypoints with confidence above 0.3, and SMPL-X parameters are optimized by minimizing the $L_1$ distance between projected and predicted 2D keypoints across all views with confidence above 0.3.
Additional regularization terms suppress unnatural head leaning and foot bending, and the results are temporally smoothed using a Savitzky–Golay filter for stability.
While existing fittings rely on triangulated 3D keypoints that are prone to triangulation errors, our pipeline directly leverages multi-view 2D keypoints, eliminating such errors and achieving higher fitting accuracy.

We manually inspected all fittings by rendering them as multi-view videos, checking both pixel-level alignment and temporal consistency.
Captures with unreliable 2D keypoints—typically from subjects wearing extremely loose clothing—are excluded from training.
In such cases, the underlying body pose is heavily occluded by the cloth, making keypoint detection difficult and resulting in noisy original SMPL-X fittings.
We also exclude captures involving human–object interactions, which are manually identified, since objects are not part of our avatar modeling.
Finally, we omit the mask loss during optimization because the SMPL-X model represents only the naked body, and using mask supervision led to body-shape distortions for subjects wearing loose garments.
\begin{figure}[t]
\begin{center}
\includegraphics[width=0.9\linewidth]{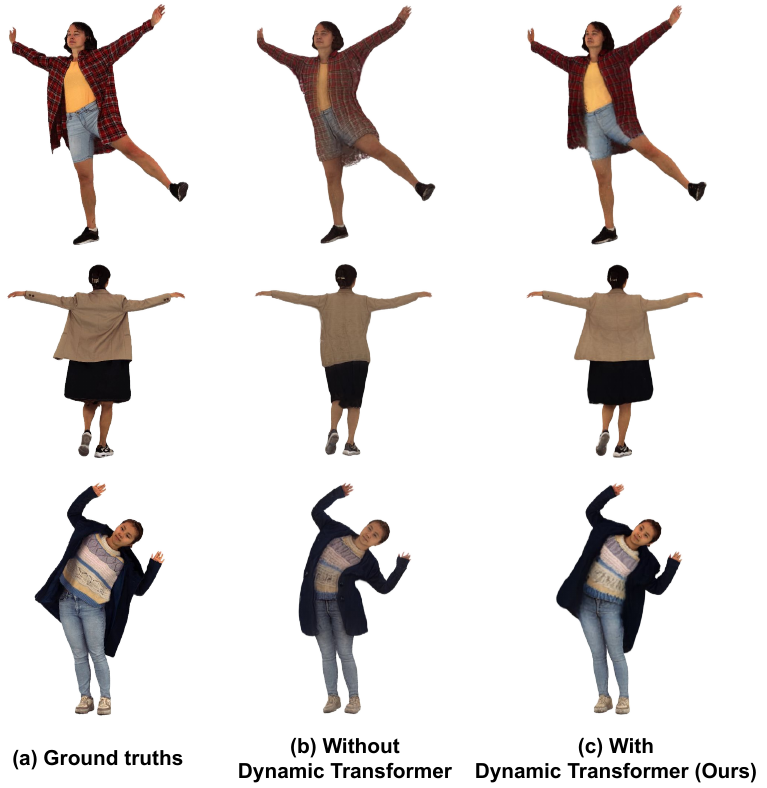}
\end{center}
\vspace*{-6mm}
\caption{
Effectiveness of our Dynamic Transformer.
}
\vspace*{-3mm}
\label{fig:ablation_dynamic_transformer}
\end{figure}

\section{Experiments}

\subsection{Datasets and evaluation metrics}

\noindent\textbf{Datasets.}
DNA-Rendering~\cite{cheng2023dna} and 4D-Dress~\cite{wang20244d} are used for training, and we evaluate on DNA-Rendering, 4D-Dress, and Actors-HQ~\cite{icsik2023humanrf}.
For DNA-Rendering, we split the dataset so that training and testing subjects do not overlap.
For 4D-Dress, we render each 3D scan from 24 uniformly placed virtual cameras, ensuring that the test set contains novel subjects and clothing.
We use all 14 sequences of Actors-HQ with the 39 visible cameras.
Actors-HQ is not used for training and therefore serves as a cross-domain generalization benchmark.
These datasets provide high-resolution videos of diverse subjects, motions, and cloth dynamics, making them well suited for assessing dynamic avatar reconstruction.
As described in Sec.~\ref{sec:reannotate}, we use our reannotated SMPL-X fittings for all experiments including baselines and previous works.

\noindent\textbf{Evaluation metrics.}
Following previous works~\cite{moon2024expressive,zhuang2025idol,qiu2025anigs,qiu2025LHM,sim2025persona,xu2025sequential}, we use PSNR, SSIM, and LPIPS~\cite{zhang2018unreasonable} as evaluation metrics, which measure the similarity between the rendered and the ground truth images.

\begin{figure}[t]
\begin{center}
\includegraphics[width=\linewidth]{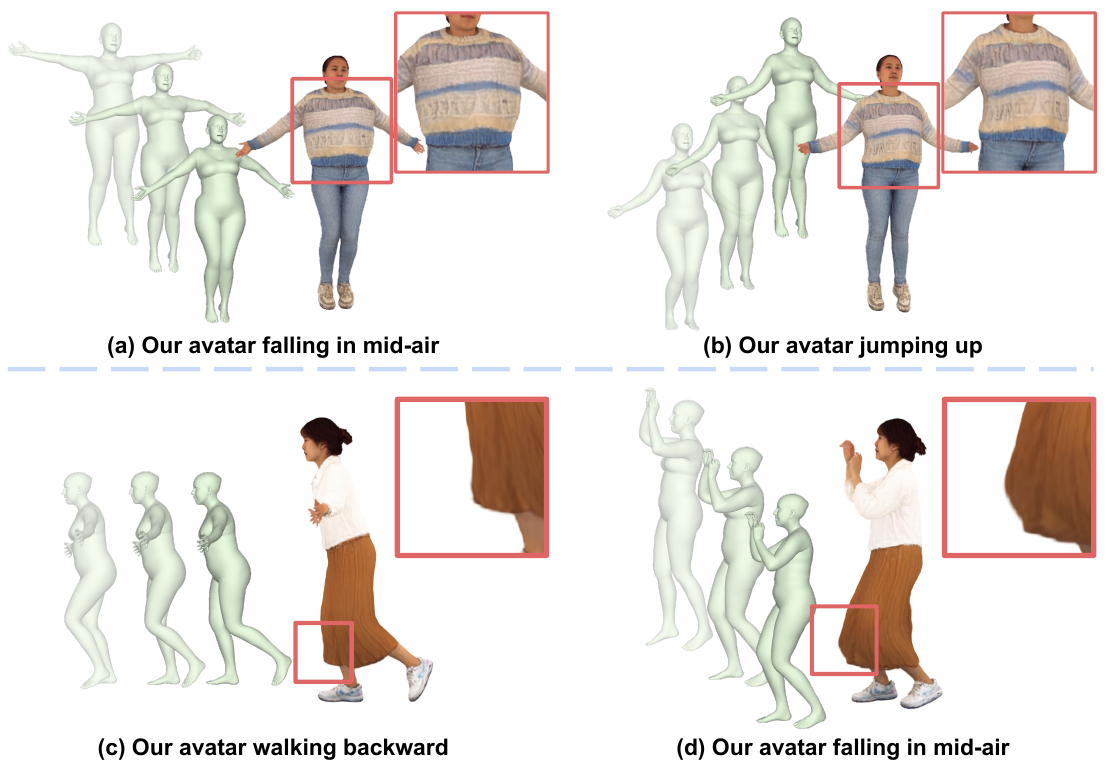}
\end{center}
\vspace*{-6mm}
\caption{
Rendered avatars in similar poses but different motion histories.
(a) and (b) share similar poses, but (a) is falling in mid-air while (b) is jumping up.
(c) and (d) also exhibit similar poses, with (c) walking backward and (d) falling in mid-air.
Despite having nearly identical poses, their clothing appears clearly different because our Dynamic Transformer effectively handles different motion information.
}
\vspace*{-3mm}
\label{fig:ablation_same_pose_diff_motion}
\end{figure}

\begin{table}[t]
\footnotesize
\centering
\setlength\tabcolsep{1.0pt}
\def\arraystretch{1.1}
\caption{
Effectiveness of our Dynamic Transformer on 4D-Dress.
}
\vspace*{-3mm}
\scalebox{1.0}{
\begin{tabular}{L{3.5cm}|C{1.2cm}C{1.2cm}C{1.2cm}}
\specialrule{.1em}{.05em}{.05em}
Settings & PSNR\textuparrow & SSIM\textuparrow & LPIPS\textdownarrow \\ \hline
Wo. Dynamic Transformer &  22.57 & 0.952 & 0.068 \\
\textbf{DynaAvatar (Ours)} & \textbf{23.62} & \textbf{0.958} & \textbf{0.062} \\ 
\specialrule{.1em}{.05em}{.05em}
\end{tabular}
}
\label{table:ablation_dynamic_transformer}
\vspace*{-3mm}
\end{table}

\begin{figure*}[t]
\begin{center}
\includegraphics[width=0.91\linewidth]{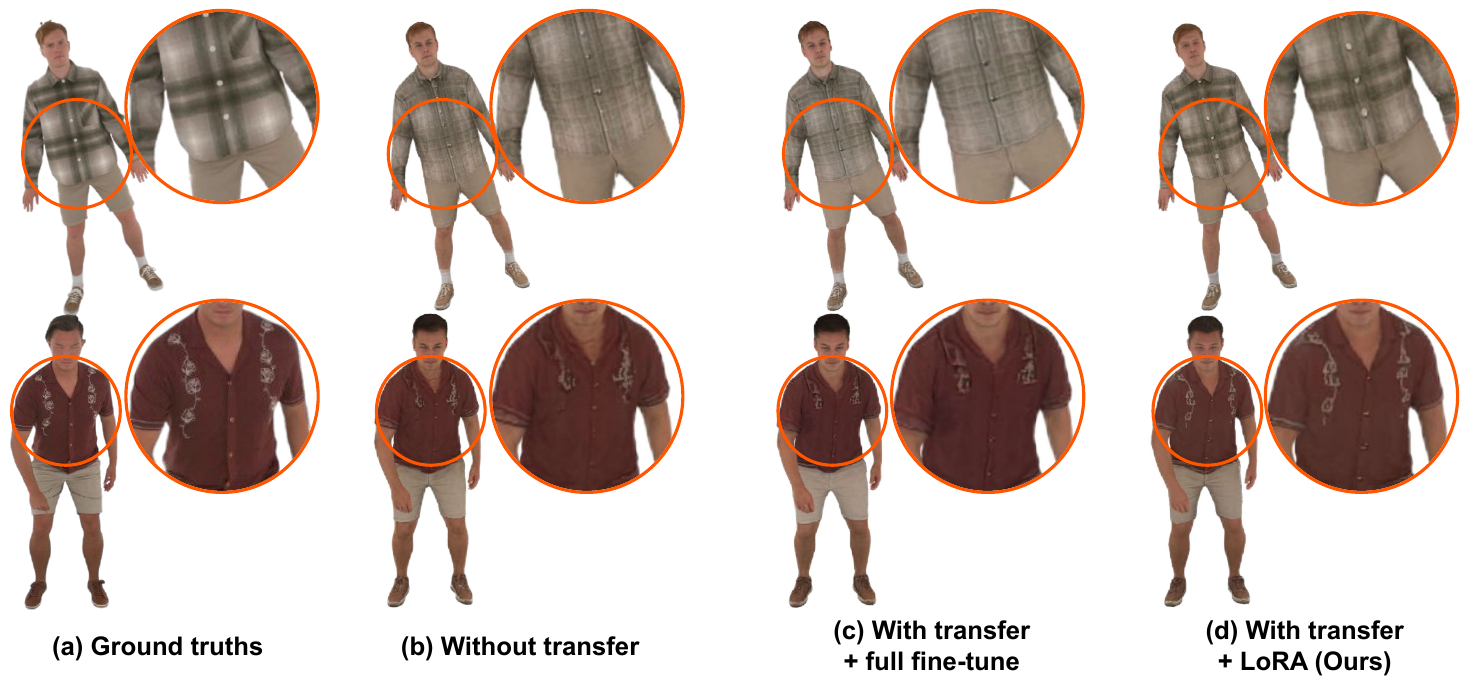}
\end{center}
\vspace*{-6mm}
\caption{
Effectiveness of our static-to-dynamic knowledge transfer with LoRA.
}
\vspace*{-3mm}
\label{fig:ablation_static_to_dynamic}
\end{figure*}

\begin{figure*}[t]
\begin{center}
\includegraphics[width=0.91\linewidth]{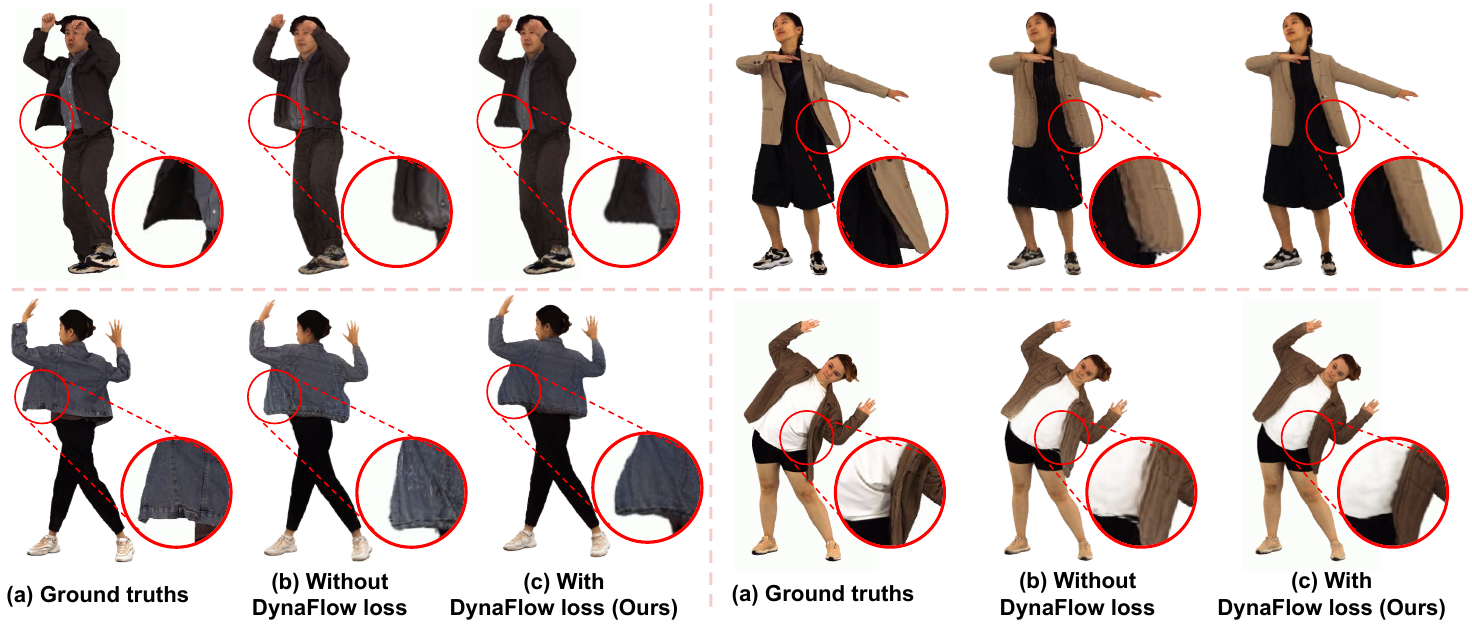}
\end{center}
\vspace*{-6mm}
\caption{
Effectiveness of our DynaFlow loss function.
}
\vspace*{-3mm}
\label{fig:ablation_dynaflow_loss}
\end{figure*}

\begin{figure*}[t]
\begin{center}
\includegraphics[width=0.95\linewidth]{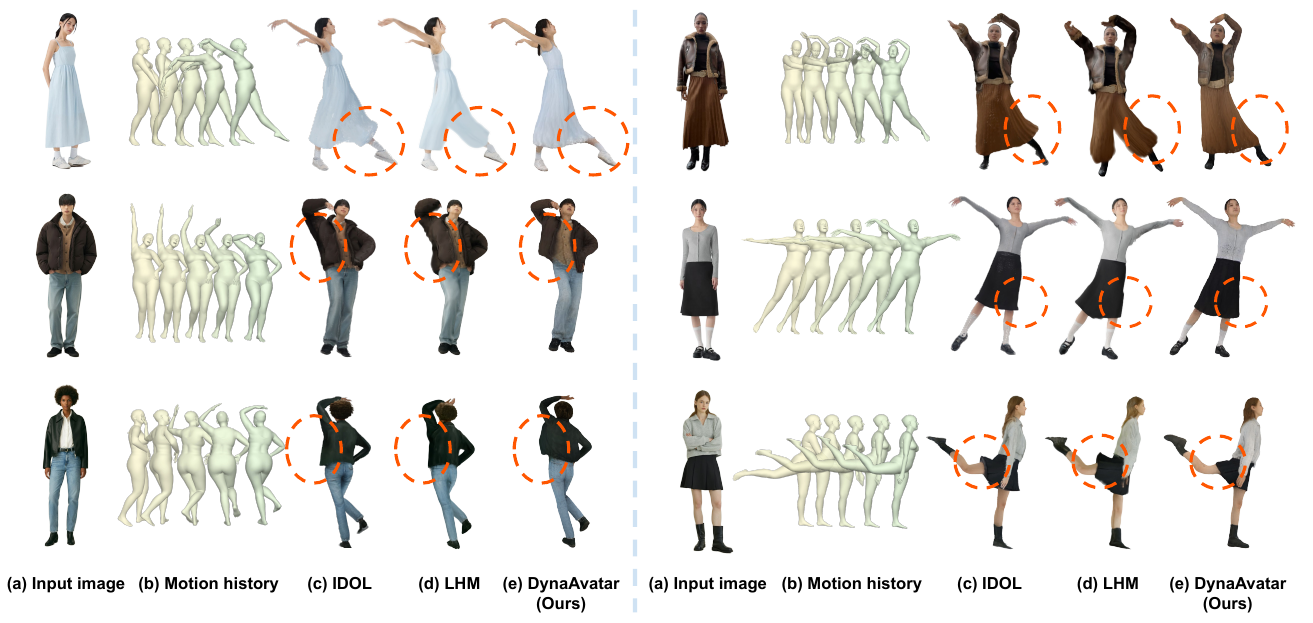}
\end{center}
\vspace*{-6mm}
\caption{
Comparison between DynaAvatar and previous single-image–based state-of-the-art methods~\cite{zhuang2025idol,qiu2025LHM} on in-the-wild images.
}
\vspace*{-3mm}
\label{fig:comparison_sota}
\end{figure*}

\begin{table*}[t]
\footnotesize
\centering
\setlength\tabcolsep{1.0pt}
\def\arraystretch{1.1}
\caption{
Comparison between DynaAvatar and previous single-image–based state-of-the-art methods.
}
\vspace*{-3mm}
\scalebox{1.0}{
\begin{tabular}{L{2.5cm}|C{1.1cm}C{1.1cm}C{1.1cm}|C{1.1cm}C{1.1cm}C{1.1cm}|C{1.1cm}C{1.1cm}C{1.1cm}}
\specialrule{.1em}{.05em}{.05em}
\multirow{2}{*}{Methods} & \multicolumn{3}{c|}{DNA-Rendering~\cite{cheng2023dna}} & \multicolumn{3}{c|}{4D-Dress~\cite{wang20244d}} & \multicolumn{3}{c}{Actors-HQ~\cite{icsik2023humanrf}} \\
& PSNR\textuparrow & SSIM\textuparrow & LPIPS\textdownarrow & PSNR\textuparrow & SSIM\textuparrow & LPIPS\textdownarrow & PSNR\textuparrow & SSIM\textuparrow & LPIPS\textdownarrow \\ \hline
IDOL~\cite{zhuang2025idol} & \cellcolor{yellow!20} 17.84 & \cellcolor{yellow!20} 0.902 & \cellcolor{yellow!20} 0.155 & \cellcolor{yellow!20} 21.31 & 0.948 & \cellcolor{yellow!20} 0.077 & \cellcolor{yellow!20} 20.93 & \cellcolor{yellow!20} 0.910 & \cellcolor{yellow!20} 0.138  \\ 
PERSONA~\cite{sim2025persona} & 14.91 & 0.883 & 0.207 & 19.46 & 0.943 & 0.098 & 19.08 & 0.904 & 0.171  \\ 
LHM~\cite{qiu2025LHM} & 17.42 & 0.901 & 0.169 & 21.03 & \cellcolor{yellow!20} 0.950 & 0.085 & 20.29 & 0.908 & 0.151  \\ 
\textbf{DynaAvatar (Ours)} & \cellcolor{red!20} \textbf{19.45} & \cellcolor{red!20} \textbf{0.916} & \cellcolor{red!20} \textbf{0.136} & \cellcolor{red!20} \textbf{23.74} & \cellcolor{red!20} \textbf{0.960} & \cellcolor{red!20} \textbf{0.064} & \cellcolor{red!20} \textbf{21.38} & \cellcolor{red!20} \textbf{0.916} & \cellcolor{red!20} \textbf{0.128} \\ 
\specialrule{.1em}{.05em}{.05em}
\end{tabular}
}
\label{table:comparison_sota}
\vspace*{-3mm}
\end{table*}

\subsection{Ablation studies}

\noindent\textbf{Dynamic Transformer.}
Fig.~\ref{fig:ablation_dynamic_transformer} and Tab.~\ref{table:ablation_dynamic_transformer} show that our Dynamic Transformer is essential for modeling motion-dependent cloth dynamics.
Fig.~\ref{fig:ablation_same_pose_diff_motion} shows that our avatars exhibit clearly different cloth deformations even though they share similar poses and originate from the same input images.
Since the only difference between the two cases is the motion history, this result demonstrates that our Dynamic Transformer effectively provides essential motion information.
A baseline without the Dynamic Transformer instead uses a Static Transformer with LoRA~\cite{hu2022lora} to extract Gaussian features, which are then fed into the Gaussian decoder.
However, because this baseline does not incorporate motion history—similar to prior single-image methods~\cite{qiu2025LHM,zhuang2025idol,qiu2025anigs}—its ability to represent dynamic deformations is fundamentally limited.

\noindent\textbf{Static-to-dynamic knowledge transfer.}
Fig.~\ref{fig:ablation_static_to_dynamic} shows that our static-to-dynamic knowledge transfer with LoRA achieves the best results.
To validate this design, we evaluate two baselines.
First ((b)), we randomly initialize the Static Transformer and train it from scratch, introducing no knowledge transfer from pretrained static models.
Second ((c)), we fully fine-tune the pretrained Static Transformer without LoRA.
As shown in the figure, both (b) and (c) fail to preserve the texture patterns of the person in the input image.
This analysis demonstrates that knowledge transfer from static captures is beneficial, and that LoRA is essential for preserving this knowledge while enabling the Dynamic Transformer to learn motion-dependent cloth dynamics.

\noindent\textbf{DynaFlow loss function.}
Fig.~\ref{fig:ablation_dynaflow_loss} shows that our DynaFlow loss function is highly effective in modeling both large cloth motions and clear cloth boundaries.
The two examples in the left column demonstrate that DynaFlow enables the model to recover large, sweeping cloth dynamics. Without DynaFlow, the model relies solely on image reconstruction losses, which operate on local patches and therefore struggle to establish accurate correspondences under large deformations—resulting in clothes that remain nearly static.
The two examples in the right column show that DynaFlow also produces noticeably cleaner cloth boundaries; in contrast, without it, the boundaries often collapse or blend into neighboring regions. This improvement stems from the geometry-only, flow-based supervision provided by DynaFlow, which removes the color–geometry ambiguity inherent to image losses. By injecting explicit pixel-displacement cues that tell each Gaussian how it should move, DynaFlow allows the model to recover both large-scale deformations and sharp, well-separated boundary details that standard image losses alone cannot supervise.

\subsection{Comparisons to state-of-the-art methods}

Fig.~\ref{fig:comparison_sota} and Tab.~\ref{table:comparison_sota} show that DynaAvatar outperforms prior single-image 3D animatable avatar methods~\cite{zhuang2025idol,sim2025persona,qiu2025LHM}.
Qualitatively, DynaAvatar models motion-dependent cloth dynamics far more faithfully, even for in-the-wild examples.

In contrast, IDOL~\cite{zhuang2025idol} and LHM~\cite{qiu2025LHM} largely preserve the cloth deformation from the input image rather than adapting it to motion.
As seen in the left-top example, the skirt is copied directly, causing unnatural ankle overlap; in the left-middle and left-bottom cases, the jacket fails to lift during upward motion; and in the right column, the skirt remains static across poses.

DynaAvatar succeeds because the Dynamic Transformer incorporates motion cues and is trained with the DynaFlow loss, which provides geometry-only, flow-based correspondence signals that resolve the color–geometry ambiguity of conventional image losses.
All previous methods are evaluated using their official implementations with our reannotated SMPL-X fittings.
For LHM, we compare against the 500M model to match our Static Transformer’s scale, and PF-LHM is excluded due to the lack of publicly available code.

\section{Conclusion}

We presented DynaAvatar, a zero-shot framework that reconstructs animatable 3D avatars with motion-dependent cloth dynamics from a single image. 
Through static-to-dynamic knowledge transfer and the proposed DynaFlow loss, our method effectively learns dynamic deformations. 
Our reannotated fittings further provide reliable supervision for training dynamic avatars. 
We believe DynaAvatar offers a promising step toward more expressive single-image avatar generation.

\section*{Acknowledgments}
This work was partly supported by the Institute of Information \& Communications Technology Planning \& Evaluation(IITP)-ICT Creative Consilience Program grant funded by the Korea government(MSIT)(IITP-2026-RS-2020-II201819, 20\%).
This research was supported by the Culture, Sports and Tourism R\&D Program through the Korea Creative Content Agency grant funded by the Ministry of Culture, Sports and Tourism in 2026(Project Name: Development of AI-based image expansion and service technology for high-resolution (8K/16K) service of performance contents, Project Number: RS-2024-00395886, Contribution Rate: 20\%).
This work was supported by the Industrial Technology Innovation Program(RS-2025-02653087, Development of a Motion Data Collection System and Dynamic Persona Modeling Technology) funded By the Ministry of Trade, Industry \& Energy(MOTIE, Korea).
This work was supported by the IITP grant funded by the MSIT (No. RS-2025-25441838, Development of a human foundation model for human-centric universal artificial intelligence and training of personnel).
This work was supported by the National Research Foundation of Korea(NRF) grant funded by the Korea government(MSIT)(RS-2025-21063115).

{
    \small
    \bibliographystyle{ieeenat_fullname}
    \bibliography{main}

@inproceedings{zhuang2025idol,
  title={{IDOL}: Instant photorealistic {3D} human creation from a single image},
  author={Zhuang, Yiyu and Lv, Jiaxi and Wen, Hao and Shuai, Qing and Zeng, Ailing and Zhu, Hao and Chen, Shifeng and Yang, Yujiu and Cao, Xun and Liu, Wei},
  booktitle={CVPR},
  year={2025}
}

@inproceedings{qiu2025anigs,
    title={{AniGS}: Animatable Gaussian Avatar from a Single Image with Inconsistent Gaussian Reconstruction},
    author={Qiu, Lingteng and Zhu, Shenhao and Zuo, Qi and Gu, Xiaodong and Dong, Yuan and Zhang, Junfei and Xu, Chao and Li, Zhe and Yuan, Weihao and Bo, Liefeng and others},
    booktitle={CVPR},
    year={2025}
}

@inproceedings{qiu2025LHM,
  title={{LHM}: Large Animatable Human Reconstruction Model from a Single Image in Seconds},
  author={Lingteng Qiu and Xiaodong Gu and Peihao Li  and Qi Zuo
     and Weichao Shen and Junfei Zhang and Kejie Qiu and Weihao Yuan
     and Guanying Chen and Zilong Dong and Liefeng Bo 
    },
  booktitle={ICCV},
  year={2025}
}

@inproceedings{moon2024expressive,
  title={Expressive whole-body {3D} gaussian avatar},
  author={Moon, Gyeongsik and Shiratori, Takaaki and Saito, Shunsuke},
  booktitle={ECCV},
  year={2024}
}

@inproceedings{sim2025persona,
  title={{PERSONA}: Personalized Whole-Body {3D} Avatar with Pose-Driven Deformations from a Single Image},
  author = {Sim, Geonhee and Moon, Gyeongsik},  
  booktitle={ICCV},
  year={2025}
}

@inproceedings{xu2025sequential,
  title={Sequential Gaussian Avatars with Hierarchical Spatio-temporal Context},
  author={Xu, Wangze and Zhan, Yifan and Zhong, Zhihang and Sun, Xiao},
  booktitle={ICCV},
  year={2025}
}

@article{qiu2025pf,
  title={{PF-LHM}: {3D} Animatable Avatar Reconstruction from Pose-free Articulated Human Images},
  author={Qiu, Lingteng and Li, Peihao and Zuo, Qi and Gu, Xiaodong and Dong, Yuan and Yuan, Weihao and Zhu, Siyu and Han, Xiaoguang and Chen, Guanying and Dong, Zilong},
  journal={arXiv preprint arXiv:2506.13766},
  year={2025}
}

@article{hu2023gaussianavatar,
  title={{GaussianAvatar}: Towards realistic human avatar modeling from a single video via animatable {3D} gaussians},
  author={Hu, Liangxiao and Zhang, Hongwen and Zhang, Yuxiang and Zhou, Boyao and Liu, Boning and Zhang, Shengping and Nie, Liqiang},
  journal={arXiv preprint arXiv:2312.02134},
  year={2023}
}

@article{bagautdinov2021driving,
  title={Driving-signal aware full-body avatars},
  author={Bagautdinov, Timur and Wu, Chenglei and Simon, Tomas and Prada, Fabian and Shiratori, Takaaki and Wei, Shih-En and Xu, Weipeng and Sheikh, Yaser and Saragih, Jason},
  journal={ACM TOG},
  year={2021}
}

@inproceedings{peng2021neural,
  title={{Neural Body}: Implicit neural representations with structured latent codes for novel view synthesis of dynamic humans},
  author={Peng, Sida and Zhang, Yuanqing and Xu, Yinghao and Wang, Qianqian and Shuai, Qing and Bao, Hujun and Zhou, Xiaowei},
  booktitle={CVPR},
  year={2021}
}

@inproceedings{peng2021animatable,
  title={Animatable neural radiance fields for modeling dynamic human bodies},
  author={Peng, Sida and Dong, Junting and Wang, Qianqian and Zhang, Shangzhan and Shuai, Qing and Zhou, Xiaowei and Bao, Hujun},
  booktitle={ICCV},
  year={2021}
}

@inproceedings{kwon2021neural,
  title={{Neural Human Performer}: Learning generalizable radiance fields for human performance rendering},
  author={Kwon, Youngjoong and Kim, Dahun and Ceylan, Duygu and Fuchs, Henry},
  booktitle={NeurIPS},
  year={2021}
}

@article{zheng2023avatarrex,
  title={{AvatarRex}: Real-time expressive full-body avatars},
  author={Zheng, Zerong and Zhao, Xiaochen and Zhang, Hongwen and Liu, Boning and Liu, Yebin},
  journal={ACM TOG},
  year={2023}
}

@inproceedings{li2024animatable,
  title={{Animatable Gaussians}: Learning pose-dependent gaussian maps for high-fidelity human avatar modeling},
  author={Li, Zhe and Zheng, Zerong and Wang, Lizhen and Liu, Yebin},
  booktitle={CVPR},
  year={2024}
}

@inproceedings{moreau2024human,
  title={Human gaussian splatting: Real-time rendering of animatable avatars},
  author={Moreau, Arthur and Song, Jifei and Dhamo, Helisa and Shaw, Richard and Zhou, Yiren and P{\'e}rez-Pellitero, Eduardo},
  booktitle={CVPR},
  year={2024}
}

@inproceedings{vaswani2017attention,
  title={Attention is all you need},
  author={Vaswani, Ashish and Shazeer, Noam and Parmar, Niki and Uszkoreit, Jakob and Jones, Llion and Gomez, Aidan N and Kaiser, {\L}ukasz and Polosukhin, Illia},
  booktitle={NeurIPS},
  year={2017}
}

@article{mildenhall2021nerf,
  title={{NeRF}: Representing scenes as neural radiance fields for view synthesis},
  author={Mildenhall, Ben and Srinivasan, Pratul P and Tancik, Matthew and Barron, Jonathan T and Ramamoorthi, Ravi and Ng, Ren},
  journal={Communications of the ACM},
  year={2021}
}

@article{kerbl20233d,
  title={{3D} Gaussian Splatting for Real-Time Radiance Field Rendering},
  author={Kerbl, Bernhard and Kopanas, Georgios and Leimk{\"u}hler, Thomas and Drettakis, George},
  journal={ACM TOG},
  year={2023}
}

@inproceedings{grigorev2023hood,
  title={{HOOD}: Hierarchical graphs for generalized modelling of clothing dynamics},
  author={Grigorev, Artur and Black, Michael J and Hilliges, Otmar},
  booktitle={CVPR},
  year={2023}
}

@article{grigorev2024contourcraft,
  title={{ContourCraft}: Learning to resolve intersections in neural multi-garment simulations},
  author={Grigorev, Artur and Becherini, Giorgio and Black, Michael and Hilliges, Otmar and Thomaszewski, Bernhard},
  journal={ACM TOG},
  year={2024}
}

@inproceedings{guo2025pgc,
  title={{PGC}: Physics-Based Gaussian Cloth from a Single Pose},
  author={Guo, Michelle and Chiang, Matt Jen-Yuan and Santesteban, Igor and Sarafianos, Nikolaos and Chen, Hsiao-yu and Halimi, Oshri and Bo{\v{z}}i{\v{c}}, Alja{\v{z}} and Saito, Shunsuke and Wu, Jiajun and Liu, C Karen and others},
  booktitle={CVPR},
  year={2025}
}

@inproceedings{lee2025mpmavatar,
  title={{MPMAvatar}: Learning {3D} gaussian avatars with accurate and robust physics-based dynamics},
  author={Lee, Changmin and Lee, Jihyun and Kim, Tae-Kyun},
  booktitle={NeurIPS},
  year={2025}
}

@inproceedings{zhan2025real,
  title={Real-time High-fidelity Gaussian Human Avatars with Position-based Interpolation of Spatially Distributed MLPs},
  author={Zhan, Youyi and Shao, Tianjia and Yang, Yin and Zhou, Kun},
  booktitle={CVPR},
  year={2025}
}

@inproceedings{rong2025gaussian,
  title={{Gaussian Garments}: Reconstructing simulation-ready clothing with photorealistic appearance from multi-view video},
  author={Rong, Boxiang and Grigorev, Artur and Wang, Wenbo and Black, Michael J and Thomaszewski, Bernhard and Tsalicoglou, Christina and Hilliges, Otmar},
  booktitle={3DV},
  year={2025}
}

@inproceedings{nakayama2025aipparel,
  title={{AIpparel}: A Multimodal Foundation Model for Digital Garments},
  author={Nakayama, Kiyohiro and Ackermann, Jan and Kesdogan, Timur Levent and Zheng, Yang and Korosteleva, Maria and Sorkine-Hornung, Olga and Guibas, Leonidas J and Yang, Guandao and Wetzstein, Gordon},
  booktitle={CVPR},
  year={2025}
}

@inproceedings{bian2025chatgarment,
  title={{ChatGarment}: Garment estimation, generation and editing via large language models},
  author={Bian, Siyuan and Xu, Chenghao and Xiu, Yuliang and Grigorev, Artur and Liu, Zhen and Lu, Cewu and Black, Michael J and Feng, Yao},
  booktitle={CVPR},
  year={2025}
}

@inproceedings{habermann2020deepcap,
  title={{DeepCap}: Monocular human performance capture using weak supervision},
  author={Habermann, Marc and Xu, Weipeng and Zollhofer, Michael and Pons-Moll, Gerard and Theobalt, Christian},
  booktitle={CVPR},
  year={2020}
}

@inproceedings{guo2024reloo,
  title={{ReLoo}: Reconstructing humans dressed in loose garments from monocular video in the wild},
  author={Guo, Chen and Jiang, Tianjian and Kaufmann, Manuel and Zheng, Chengwei and Valentin, Julien and Song, Jie and Hilliges, Otmar},
  booktitle={ECCV},
  year={2024}
}

@inproceedings{qiu2023rec,
  title={{REC-MV}: Reconstructing {3D} dynamic cloth from monocular videos},
  author={Qiu, Lingteng and Chen, Guanying and Zhou, Jiapeng and Xu, Mutian and Wang, Junle and Han, Xiaoguang},
  booktitle={CVPR},
  year={2023}
}

@inproceedings{chen2024within,
  title={Within the dynamic context: Inertia-aware {3D} human modeling with pose sequence},
  author={Chen, Yutong and Zhan, Yifan and Zhong, Zhihang and Wang, Wei and Sun, Xiao and Qiao, Yu and Zheng, Yinqiang},
  booktitle={ECCV},
  year={2024}
}

@inproceedings{hu2022lora,
  title={{LoRA}: Low-rank adaptation of large language models.},
  author={Hu, Edward J and Shen, Yelong and Wallis, Phillip and Allen-Zhu, Zeyuan and Li, Yuanzhi and Wang, Shean and Wang, Lu and Chen, Weizhu and others},
  booktitle={ICLR},
  year={2022}
}

@inproceedings{cheng2023dna,
  title={{DNA-Rendering}: A diverse neural actor repository for high-fidelity human-centric rendering},
  author={Cheng, Wei and Chen, Ruixiang and Fan, Siming and Yin, Wanqi and Chen, Keyu and Cai, Zhongang and Wang, Jingbo and Gao, Yang and Yu, Zhengming and Lin, Zhengyu and others},
  booktitle={ICCV},
  year={2023}
}

@inproceedings{wang20244d,
  title={{4D-DRESS}: A {4D} dataset of real-world human clothing with semantic annotations},
  author={Wang, Wenbo and Ho, Hsuan-I and Guo, Chen and Rong, Boxiang and Grigorev, Artur and Song, Jie and Zarate, Juan Jose and Hilliges, Otmar},
  booktitle={CVPR},
  year={2024}
}

@inproceedings{han2023high,
  title={High-fidelity {3D} human digitization from single 2k resolution images},
  author={Han, Sang-Hun and Park, Min-Gyu and Yoon, Ju Hong and Kang, Ju-Mi and Park, Young-Jae and Jeon, Hae-Gon},
  booktitle={CVPR},
  year={2023}
}

@inproceedings{yu2021function4d,
  title={{Function4D}: Real-time human volumetric capture from very sparse consumer rgbd sensors},
  author={Yu, Tao and Zheng, Zerong and Guo, Kaiwen and Liu, Pengpeng and Dai, Qionghai and Liu, Yebin},
  booktitle={CVPR},
  year={2021}
}

@inproceedings{liu2016deepfashion,
  title={{DeepFashion}: Powering robust clothes recognition and retrieval with rich annotations},
  author={Liu, Ziwei and Luo, Ping and Qiu, Shi and Wang, Xiaogang and Tang, Xiaoou},
  booktitle={CVPR},
  year={2016}
}

@misc{renderpeople,
  author       = {RenderPeople},
  howpublished = {\\url{https://renderpeople.com}},
  year         = {2025}
}

@inproceedings{xiong2024mvhumannet,
  title={{MVHumanNet}: A large-scale dataset of multi-view daily dressing human captures},
  author={Xiong, Zhangyang and Li, Chenghong and Liu, Kenkun and Liao, Hongjie and Hu, Jianqiao and Zhu, Junyi and Ning, Shuliang and Qiu, Lingteng and Wang, Chongjie and Wang, Shijie and others},
  booktitle={CVPR},
  year={2024}
}

@article{icsik2023humanrf,
  title={{HumanRF}: High-fidelity neural radiance fields for humans in motion},
  author={I{\c{s}}{\i}k, Mustafa and R{\"u}nz, Martin and Georgopoulos, Markos and Khakhulin, Taras and Starck, Jonathan and Agapito, Lourdes and Nie{\ss}ner, Matthias},
  journal={ACM TOG},
  year={2023}
}

@article{shao2024360,
  title={360-degree human video generation with {4D} diffusion transformer},
  author={Shao, Ruizhi and Pang, Youxin and Zheng, Zerong and Sun, Jingxiang and Liu, Yebin},
  journal={ACM TOG},
  year={2024}
}

@inproceedings{xu2024magicanimate,
  title={{MagicAnimate}: Temporally consistent human image animation using diffusion model},
  author={Xu, Zhongcong and Zhang, Jianfeng and Liew, Jun Hao and Yan, Hanshu and Liu, Jia-Wei and Zhang, Chenxu and Feng, Jiashi and Shou, Mike Zheng},
  booktitle={CVPR},
  year={2024}
}

@inproceedings{hu2024animate,
  title={{Animate Anyone}: Consistent and controllable image-to-video synthesis for character animation},
  author={Li Hu and Xin Gao and Peng Zhang and Ke Sun and Bang Zhang and Liefeng Bo},
  booktitle={CVPR},
  year={2024}
}

@inproceedings{zhang2025mimicmotion,
  title={{MimicMotion}: High-Quality Human Motion Video Generation with Confidence-aware Pose Guidance},
  author={Yuang Zhang and Jiaxi Gu and Li-Wen Wang and Han Wang and Junqi Cheng and Yuefeng Zhu and Fangyuan Zou},
  booktitle={ICML},
  year={2025}
}

@inproceedings{tu2025stableanimator,
  title={{StableAnimator}: High-quality identity-preserving human image animation},
  author={Tu, Shuyuan and Xing, Zhen and Han, Xintong and Cheng, Zhi-Qi and Dai, Qi and Luo, Chong and Wu, Zuxuan},
  booktitle={CVPR},
  year={2025}
}

@inproceedings{zhu2024champ,
  title={Champ: Controllable and consistent human image animation with {3D} parametric guidance},
  author={Zhu, Shenhao and Chen, Junming Leo and Dai, Zuozhuo and Dong, Zilong and Xu, Yinghui and Cao, Xun and Yao, Yao and Zhu, Hao and Zhu, Siyu},
  booktitle={ECCV},
  year={2024}
}

@inproceedings{men2025mimo,
  title={{MIMO}: Controllable character video synthesis with spatial decomposed modeling},
  author={Men, Yifang and Yao, Yuan and Cui, Miaomiao and Bo, Liefeng},
  booktitle={CVPR},
  year={2025}
}

@inproceedings{luo2025dreamactor,
  title={{DreamActor-M1}: Holistic, expressive and robust human image animation with hybrid guidance},
  author={Luo, Yuxuan and Rong, Zhengkun and Wang, Lizhen and Zhang, Longhao and Hu, Tianshu},
  booktitle={ICCV},
  year={2025}
}

@article{hu2025humangif,
  title={{HumanGif}: Single-view human diffusion with generative prior},
  author={Hu, Shoukang and Narihira, Takuya and Fukuda, Kazumi and Sawata, Ryosuke and Shibuya, Takashi and Mitsufuji, Yuki},
  journal={arXiv preprint arXiv:2502.12080},
  year={2025}
}

@article{song2025x,
  title={{X-UniMotion}: Animating Human Images with Expressive, Unified and Identity-Agnostic Motion Latents},
  author={Song, Guoxian and Xu, Hongyi and Zhao, Xiaochen and Xie, You and Gu, Tianpei and Li, Zenan and Zhang, Chenxu and Luo, Linjie},
  journal={arXiv preprint arXiv:2508.09383},
  year={2025}
}

@inproceedings{ho2020denoising,
  title={Denoising diffusion probabilistic models},
  author={Ho, Jonathan and Jain, Ajay and Abbeel, Pieter},
  booktitle={NeurIPS},
  year={2020}
}

@article{blattmann2023stable,
  title={{Stable Video Diffusion}: Scaling latent video diffusion models to large datasets},
  author={Blattmann, Andreas and Dockhorn, Tim and Kulal, Sumith and Mendelevitch, Daniel and Kilian, Maciej and Lorenz, Dominik and Levi, Yam and English, Zion and Voleti, Vikram and Letts, Adam and others},
  journal={arXiv preprint arXiv:2311.15127},
  year={2023}
}

@inproceedings{pavlakos2019expressive,
  title={Expressive body capture: {3D} hands, face, and body from a single image},
  author={Pavlakos, Georgios and Choutas, Vasileios and Ghorbani, Nima and Bolkart, Timo and Osman, Ahmed AA and Tzionas, Dimitrios and Black, Michael J},
  booktitle={CVPR},
  year={2019}
}

@article{oquab2023dinov2,
  title={{DINOv2}: Learning robust visual features without supervision},
  author={Oquab, Maxime and Darcet, Timoth{\'e}e and Moutakanni, Th{\'e}o and Vo, Huy and Szafraniec, Marc and Khalidov, Vasil and Fernandez, Pierre and Haziza, Daniel and Massa, Francisco and El-Nouby, Alaaeldin and others},
  journal={arXiv preprint arXiv:2304.07193},
  year={2023}
}

@inproceedings{khirodkar2024sapiens,
  title={Sapiens: Foundation for human vision models},
  author={Khirodkar, Rawal and Bagautdinov, Timur and Martinez, Julieta and Zhaoen, Su and James, Austin and Selednik, Peter and Anderson, Stuart and Saito, Shunsuke},
  booktitle={ECCV},
  year={2024}
}

@inproceedings{zhang2018unreasonable,
  title={The unreasonable effectiveness of deep features as a perceptual metric},
  author={Zhang, Richard and Isola, Phillip and Efros, Alexei A and Shechtman, Eli and Wang, Oliver},
  booktitle={CVPR},
  year={2018}
}

@inproceedings{esser2024scaling,
  title={Scaling rectified flow transformers for high-resolution image synthesis},
  author={Esser, Patrick and Kulal, Sumith and Blattmann, Andreas and Entezari, Rahim and M{\"u}ller, Jonas and Saini, Harry and Levi, Yam and Lorenz, Dominik and Sauer, Axel and Boesel, Frederic and others},
  booktitle={ICML},
  year={2024}
}

@inproceedings{zhou2019continuity,
  title={On the continuity of rotation representations in neural networks},
  author={Zhou, Yi and Barnes, Connelly and Lu, Jingwan and Yang, Jimei and Li, Hao},
  booktitle={CVPR},
  year={2019}
}

@inproceedings{yang2023effective,
  title={Effective whole-body pose estimation with two-stages distillation},
  author={Yang, Zhendong and Zeng, Ailing and Yuan, Chun and Li, Yu},
  booktitle={ICCVW},
  year={2023}
}

@article{yin2025smplest,
  title={{SMPLest-X}: Ultimate scaling for expressive human pose and shape estimation},
  author={Yin, Wanqi and Cai, Zhongang and Wang, Ruisi and Zeng, Ailing and Wei, Chen and Sun, Qingping and Mei, Haiyi and Wang, Yanjun and Pang, Hui En and Zhang, Mingyuan and others},
  journal={arXiv preprint arXiv:2501.09782},
  year={2025}
}

@inproceedings{lindenberger2023lightglue,
  title={{LightGlue}: Local feature matching at light speed},
  author={Lindenberger, Philipp and Sarlin, Paul-Edouard and Pollefeys, Marc},
  booktitle={ICCV},
  year={2023}
}

@inproceedings{deng2019arcface,
  title={ArcFace: Additive Angular Margin Loss for Deep Face Recognition},
  author={Deng, Jiankang and Guo, Jia and Xue, Niannan and Zafeiriou, Stefanos},
  booktitle={Proceedings of the IEEE/CVF Conference on Computer Vision and Pattern Recognition (CVPR)},
  pages={4690--4699},
  year={2019}
}
}

\clearpage



\twocolumn[
\begin{center}
\textbf{\large Supplementary Material \textit{for} \\ \vspace{2mm}
\large{``Zero-Shot Reconstruction of Animatable 3D Avatars with Cloth Dynamics \\from a Single Image"}}

\author{
    Joohyun Kwon \qquad
    Geonhee Sim \qquad
    Gyeongsik Moon \\
    Korea University \\
    {\tt\small \{juheanqueen, kh6362, mks0601\}@korea.ac.kr} \\
    {\small \url{https://juhyeon-kwon.github.io/DynaAvatar.github.io/}}
}
\end{center}
]
\setcounter{section}{0}
\setcounter{table}{0}
\setcounter{figure}{0}
\renewcommand{\thesection}{S\arabic{section}}   
\renewcommand{\thetable}{S\arabic{table}}   
\renewcommand{\thefigure}{S\arabic{figure}}

In this supplementary material, we provide more experiments, discussions, and other details that could not be included in the main text due to the lack of pages.
The contents are summarized below:
\begin{itemize}
    \item Sec.~\ref{sec:qualitative_comparisons_suppl}: Comparisons to related works
    \item Sec.~\ref{sec:reannotation_compare_suppl}: Dataset reannotation comparisons
    \item Sec.~\ref{sec:ablation_studies_suppl}: Ablation studies
    \item Sec.~\ref{sec:architecture_details_suppl}: Architecture details
    \item Sec.~\ref{sec:implementation_details}: Implementation details
\end{itemize}


\section{Comparisons to related works}\label{sec:qualitative_comparisons_suppl}

We compare DynaAvatar with state-of-the-art methods, physics-based approaches, and diffusion-based approaches to show its advantages. Please refer to the accompanying supplementary video for the full animation results.

\begin{figure*}[t]
\begin{center}
\includegraphics[width=\linewidth]{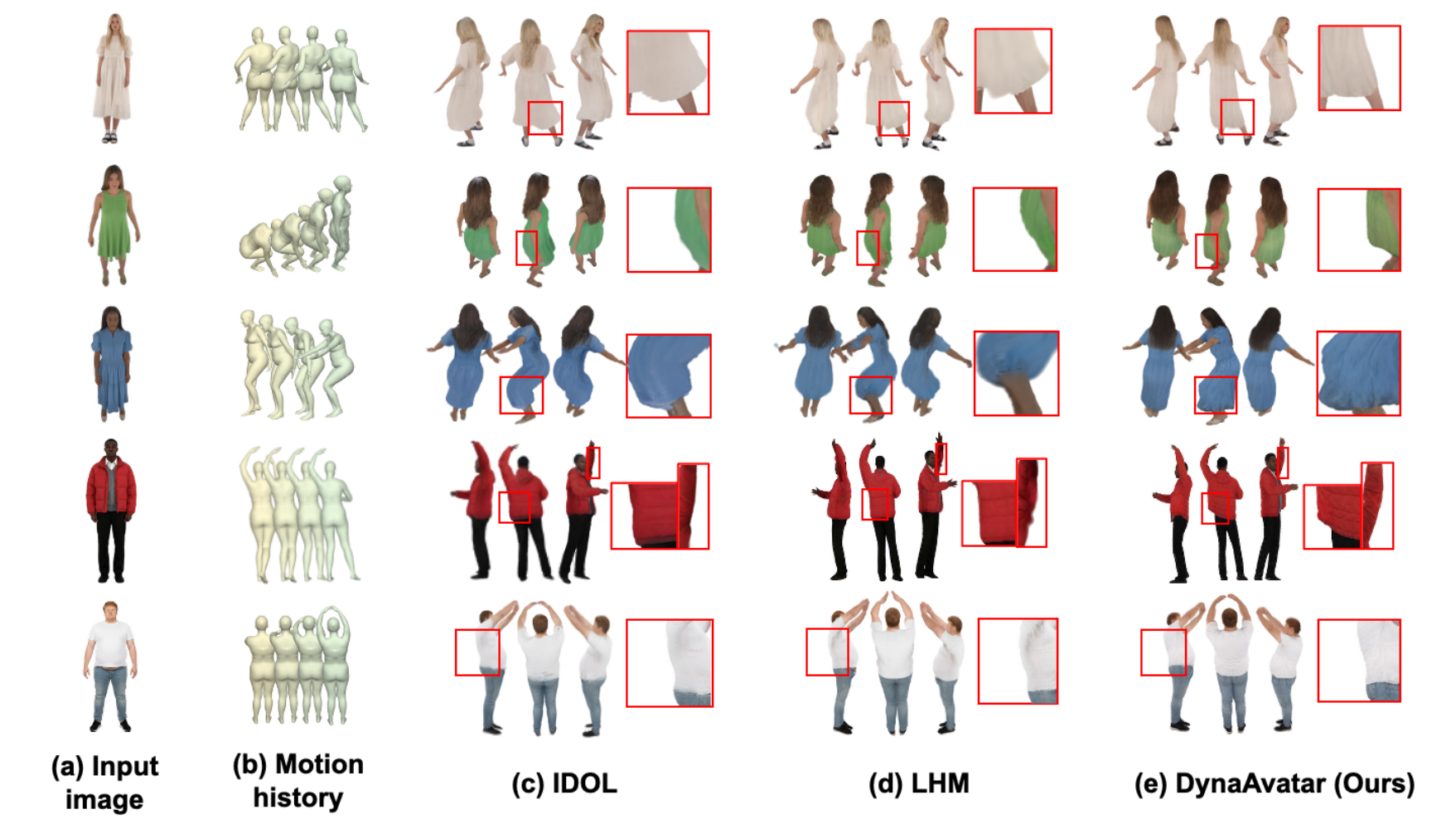}
\end{center}
\caption{
Comparison between DynaAvatar and previous single-image–based state-of-the-art methods ~\cite{qiu2025LHM, zhuang2025idol} on in-the-wild images.
}
\label{fig:sota_comp}
\end{figure*}
\subsection{Comparison to state-of-the-art methods}
\subsubsection{Qualitative comparisons}
Fig.~\ref{fig:sota_comp} shows comparisons of our DynaAvatar and previous state-of-the-art methods ~\cite{sim2025persona, zhuang2025idol, qiu2025LHM} from in-the-wild input image.
Our method successfully reconstructs and animates avatars with high-fidelity cloth dynamics.
For instance, when the subject raises their arms, the upper garment naturally lifts upward, exhibiting physically plausible motion-dependent dynamics.

In contrast, baseline methods generate animations without incorporating motion-dependent dynamics.
Consequently, the resulting animations often lack realism, as the garments remain static regardless of the body's movement.
Our method effectively overcomes this limitation by leveraging the Dynamic Transformer, resulting in superior visual realism.

Note that PF-LHM~\cite{qiu2025pf} is excluded due to code unavailability. Nevertheless, as it takes only the pose cues without motion information (\emph{i.e.}, a sequence of poses), similar to PERSONA~\cite{sim2025persona}, it is expected to lack the capability to represent motion-dependent dynamics, thereby underperforming compared to our motion-aware approach.

\begin{table}[t] 
\centering
\caption{Comparison of face consistency (FC) on DNA-Rendering.}
\small 
\begin{tabular}{l|c}
\specialrule{.1em}{.05em}{.05em}
Methods & FC $\uparrow$ \\ \hline
IDOL [\textcolor{cvprblue}{60}]    & 0.625 \\ 
LHM [\textcolor{cvprblue}{37}]     & 0.697 \\ 
\textbf{DynaAvatar (Ours)}         & \textbf{0.712} \\ 
\specialrule{.1em}{.05em}{.05em}
\end{tabular}
\label{table:fc}
\end{table}

\begin{table}[t]
\centering
\caption{Comparison of computational costs.}
\small 
\begin{tabular}{l|cc|cc}
\specialrule{.1em}{.05em}{.05em}
\multirow{2}{*}{Methods} & Zero- & Cloth & \multirow{2}{*}{Time} & \# of \\
                         & shot & dynamics &                       & params. \\ \hline
PERSONA [\textcolor{cvprblue}{43}] & \xmark & \cmark & 3.11h & 4M \\ 
LHM-500M [\textcolor{cvprblue}{37}]     & \cmark & \xmark & 1.08s & 356M \\ 
LHM-1B [\textcolor{cvprblue}{37}]       & \cmark & \xmark & 3.00s & 1.1B \\ 
\textbf{DynaAvatar (Ours)}         & \cmark & \cmark & 1.82s & 719M \\ 
\specialrule{.1em}{.05em}{.05em}
\end{tabular}
\label{table:latency}
\end{table}

\subsubsection{Quantitative comparisons}
\paragraph{Face consistency.}
Table.~\ref{table:fc} on DNA-Rendering shows that DynaAvatar achieves superior image consistency, as shown by the higher face consistency (FC) compared to baseline methods. FC is measured via cosine similarity in the ArcFace~\cite{deng2019arcface} embedding space.
\paragraph{Inference Latency.}
As shown in Table.~\ref{table:latency}, DynaAvatar ensures fast inference via its zero-shot architecture, avoiding the lengthy per-subject optimization of PERSONA. 
While our Dynamic Transformer adds moderate overhead compared to LHM-500M, it remains more efficient than LHM-1B in both time and parameters. 
This cost is essential for capturing dynamic deformations, offering a superior trade-off for motion-dependent cloth dynamics that LHM lacks.
All metrics are measured on a RTX pro 6000 GPU.

\begin{figure}[t]
\begin{center}
\includegraphics[width=\linewidth]{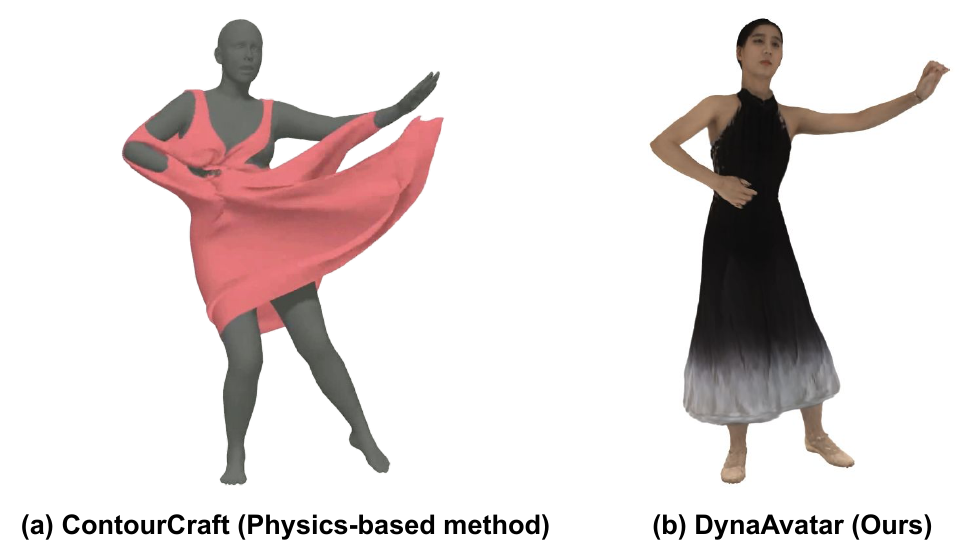}
\end{center}
\caption{
Comparison between physics-based method~\cite{grigorev2024contourcraft} and DynaAvatar.
}
\label{fig:phys_comp}
\end{figure}
\subsection{Comparison to physics-based approaches}

Fig.~\ref{fig:phys_comp} compares the physics-based method~\cite{grigorev2023hood, grigorev2024contourcraft} with DynaAvatar, highlighting the instability of the former under in-the-wild scenarios.
Note that we used a garment template of a similar type to the input image for the physics-based simulation.
As shown in the Fig.~\ref{fig:phys_comp} left, applying physics simulation to in-the-wild sequences often leads to catastrophic failures where the cloth unrealistically flies away or drifts. This instability stems from the imperfect in-the-wild pose estimation. Specifically, pose errors frequently cause the body mesh to penetrate the garment mesh, creating invalid collision constraints. These interpenetrations trigger erroneous inputs in the simulation, causing garment to become unstable. 
Moreover, these methods primarily focus on geometric garment deformation, often lacking the capability to model photorealistic, full-body appearance.

In contrast, DynaAvatar robustly synthesizes both motion-dependent cloth dynamics and high-fidelity appearance, even when driven by in-the-wild motion sequences.
These results validate DynaAvatar as a robust and practical method for animating avatars from single images.

\begin{figure}[t]
\begin{center}
\includegraphics[width=\linewidth]{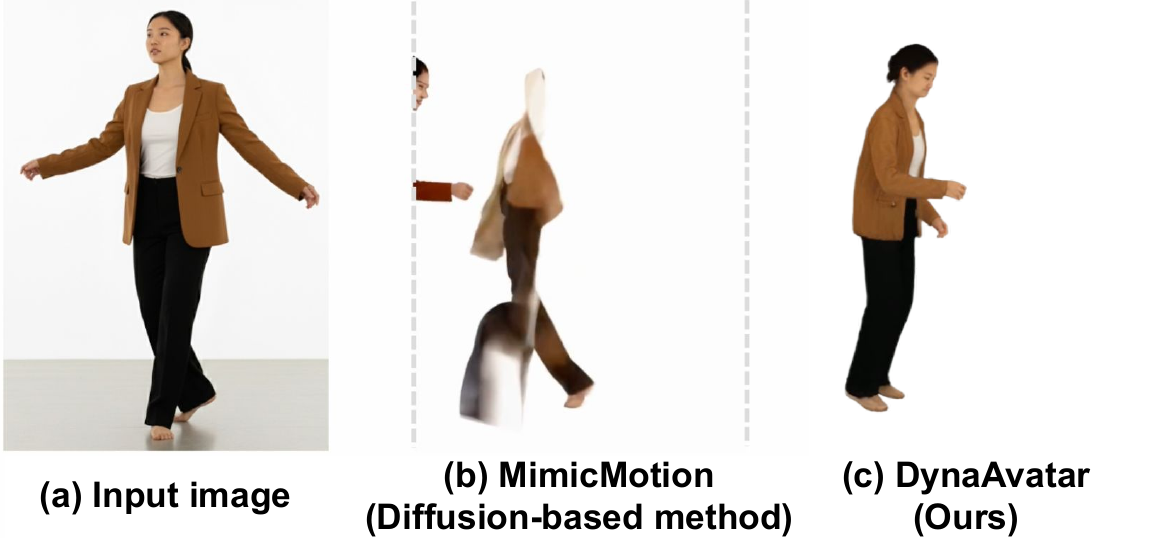}
\end{center}
\caption{
Comparison between diffusion-based method~\cite{zhang2025mimicmotion} and DynaAvatar.
}
\label{fig:diff_comp}
\end{figure}
\subsection{Comparison to diffusion-based approaches}

Fig.~\ref{fig:diff_comp} compares the state-of-the-art diffusion-based method~\cite{zhang2025mimicmotion} with DynaAvatar, highlighting the limitations of diffusion models. These models fundamentally require pixel-level alignment between the reference image and the target pose. When this constraint is violated due to large global motion, the generated results suffer from severe degradation, exhibiting noticeable artifacts and hallucinations.

Moreover, since diffusion-based methods predominantly center the target pose along the $y$-axis within a fixed output resolution, body parts such as arms are frequently cropped. Furthermore, significant movement along the $x$-axis often causes the subject to move out of frame, cutting off parts of the animation.

In contrast, DynaAvatar is free from these alignment constraints and robustly handles large global motions. This capability stems from our Dynamic Transformer, which effectively incorporates motion features via attention mechanisms without relying on explicit spatial alignment.

\definecolor{dnared}{RGB}{224,102,102}
\definecolor{hqyellow}{RGB}{255,217,102}

\begin{figure*}[t]
\begin{center}
\includegraphics[width=\linewidth]{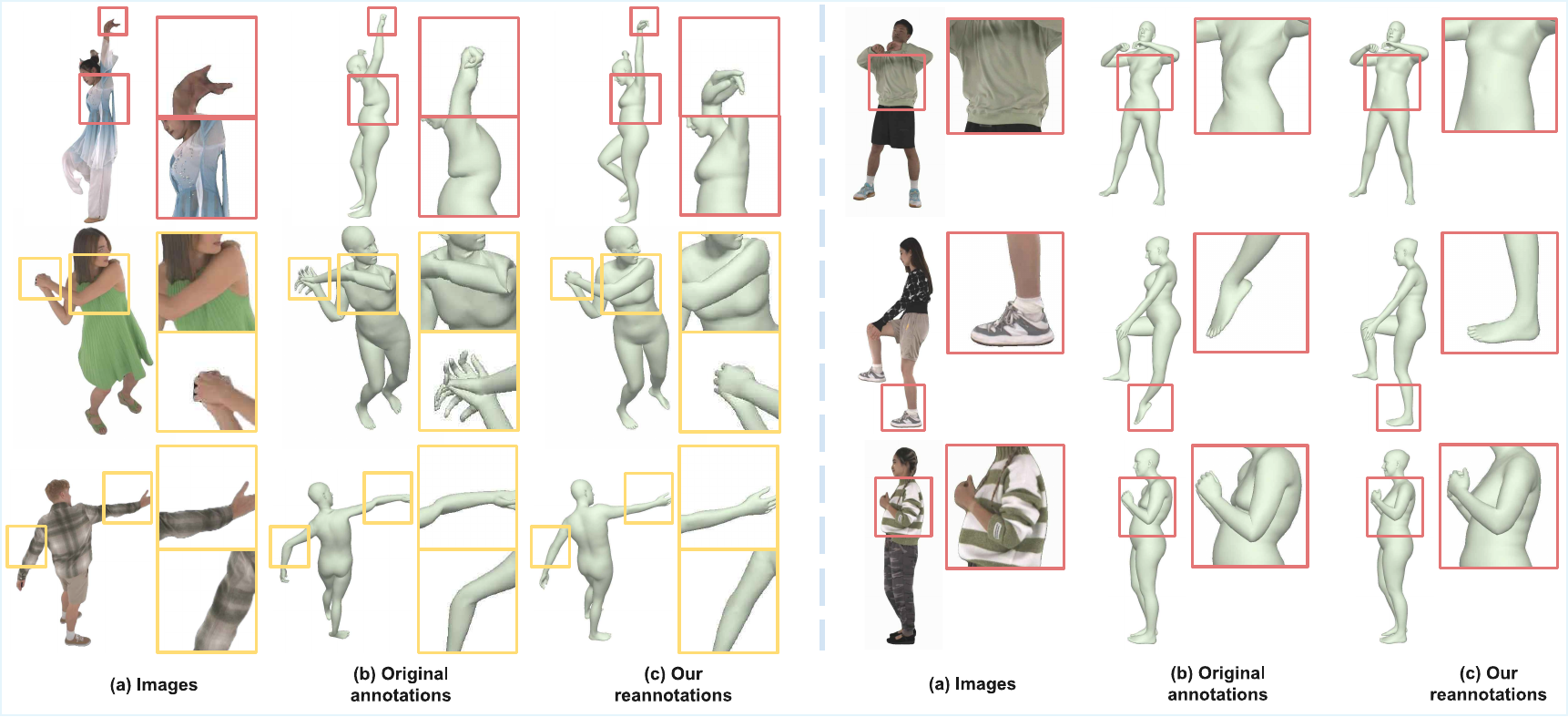}
\end{center}
\caption{
Comparison between (b) the original annotations and (c) our reannotations. The bounding box colors indicate the source datasets: \textcolor{dnared}{Red} denotes DNA-Rendering~\cite{cheng2023dna}, and \textcolor{hqyellow}{Yellow} denotes Actors-HQ~\cite{icsik2023humanrf}.
}
\label{fig:reannot_comp}
\end{figure*}


\section{Dataset Reannotation comparisons}\label{sec:reannotation_compare_suppl}
Fig.~\ref{fig:reannot_comp} provides additional comparisons between (b) the original SMPL-X fittings and (c) our reannotated results. Our reannotations produce more accurate and visually plausible poses, whereas the original annotations often contain noisy predictions and noticeable temporal jitter. Such instability in the original annotations hinders learning a reliable and practical relationship between human motion and cloth deformation. In contrast, our reannotated sequences exhibit significantly improved temporal consistency and pose accuracy. As a result, our reannotated datasets are directly usable for training motion-dependent deformation models.


\section{Ablation studies}\label{sec:ablation_studies_suppl}
We provide additional ablation study results to validate our design choices.

\begin{table}[t] 
\centering
\caption{Effectiveness of our dataset reannotations on 4D-Dress.}
\small %
\begin{tabular}{l|ccc}
\specialrule{.1em}{.05em}{.05em}
Settings & PSNR $\uparrow$ & SSIM $\uparrow$ & LPIPS $\downarrow$ \\ \hline
(1) & 20.72 & 0.952 & 0.085 \\ 
(2) & 20.91 & 0.953 & 0.084 \\ 
\textbf{(3) (Ours)} & \textbf{23.74} & \textbf{0.960} & \textbf{0.064} \\ 
\specialrule{.1em}{.05em}{.05em}
\end{tabular}
\label{table:ablation_reannotation}
\end{table}
\subsection{Dataset reannotations}
Table~\ref{table:ablation_reannotation} on 4D-Dress shows the value of our reannotations by fixing the architecture while varying training annotations. 
We compare three settings: (1) the original annotations, (2) reannotation of the originally available frames, and (3) our fully reannotated dataset (Sec.~\textcolor{cvprblue}{4}).
Results show that our reannotations (3) yield far superior results compared to the original annotations (1), which suffers from 80\% missing frames. 
Fig. \textcolor{cvprblue}{4} and Sec. \textcolor{cvprblue}{S2} additionally show the value of our reannotations.

\section{Architecture details}\label{sec:architecture_details_suppl}

Fig.~\textcolor{cvprblue}{2} and Sec.~\textcolor{cvprblue}{3.1} of the main manuscript show architecture of the proposed DynaAvatar.
We provide detailed descriptions of each component.

\subsection{Static Transformer}

The Static Transformer takes two distinct image tokens: body tokens and head tokens, extracted via Sapiens~\cite{khirodkar2024sapiens} and DINOv2~\cite{oquab2023dinov2}, respectively.
It consists of several layers, each of which is composed of a Body Transformer block and a Head Transformer block.
The Body Transformer block utilizes the body tokens as key and value to update the input query tokens, whereas the Head Transformer block utilizes the head tokens.
Additionally, we compute the global average of the body tokens and inject this feature into the Static Transformer through adaptive Layer Normalization (AdaLN)~\cite{Peebles2022DiT}.

\subsection{Motion encoder}
The Motion encoder is designed as a simple MLP that takes the motion history as input and outputs motion tokens. We construct a motion history representation from the pose sequence, which consists of $K=22$ body joints. For a pose sequence with $T$ frames, we concatenate the 3D joint linear velocities, 6D rotation-parameterized~\cite{zhou2019continuity} pose, pose velocity, and pose acceleration, resulting in a 21-dimensional motion vector per joint. This motion history is flattened to form a tensor of shape $\mathbb{R}^{T \times (K \cdot 21)}$, which is subsequently mapped to $T$ motion tokens via the motion encoder.

\subsection{Dynamic Transformer}
The Dynamic Transformer utilizes the encoded motion tokens to update the query features output by the Static Transformer.
Unlike the Static Transformer layer, which comprises two distinct blocks, Dynamic Transformer layer is implemented as a single block where motion tokens act as keys and values. 
Furthermore, the last element of the motion tokens is injected via AdaLN to explicitly provide the target pose context.


\section{Implementation details}\label{sec:implementation_details}

We observed that constructing a batch with a single subject across multiple timeframes and views yields better convergence than stacking multiple subjects.
Accordingly, for the training of DynaAvatar, we configure the batch with $F=4$ temporal frames and $V=4$ views per subject, resulting in a batch size of 16.
Our model is trained using the AdamW optimizer with an initial learning rate of $4 \times 10^{-4}$ and gradient clipping set to 0.1.
We apply LoRA to all linear layers in the Static Transformer with a rank $r=32$, scaling alpha $\alpha=64$, and a dropout rate of $0.1$.
The training is conducted on 8 NVIDIA RTX Pro 6000 GPUs for a total of 40K iterations, taking approximately 90 hours. The DynaFlow loss is activated after 20K iterations to ensure that a coarse geometry is established.


\end{document}